\documentclass[twoside,11pt]{article}

\usepackage{jmlr2e}
\usepackage{times}
\usepackage{epsfig}
\usepackage{graphicx}
\usepackage{amssymb}
\usepackage[usenames,dvipsnames,svgnames,table]{xcolor}
\graphicspath{{./}{./images/}}

\usepackage[ruled,vlined]{algorithm2e}
\usepackage{booktabs,balance,adjustbox,rotating,array,enumitem,mathtools}
\usepackage{xspace,url,subcaption,multirow}
\usepackage[backref,colorlinks=true,linkcolor=blue,citecolor=blue,urlcolor=blue]{hyperref}
\usepackage{soul}

\usepackage{commath,xspace,amssymb,mathtools,rotating}
\usepackage[backref,colorlinks=true,linkcolor=blue,citecolor=blue,urlcolor=blue]{hyperref}
\newcolumntype{L}[1]{>{\raggedright\let\newline\\\arraybackslash\hspace{0pt}}m{#1}}
\newcolumntype{C}[1]{>{\centering\let\newline\\\arraybackslash\hspace{0pt}}m{#1}}
\newcolumntype{R}[1]{>{\raggedleft\let\newline\\\arraybackslash\hspace{0pt}}m{#1}}

\usepackage{enumitem}

\graphicspath{{./}{./figures/}}

\makeatletter
\DeclareRobustCommand\onedot{\futurelet\@let@token\@onedot}
\def\@onedot{\ifx\@let@token.\else.\null\fi\xspace}

\def\eg{\emph{e.g}\onedot} 
\def\ie{\emph{i.e}\onedot}

\def\etal{\emph{et al}\onedot}
\makeatother

\def\Vec#1{{\boldsymbol{#1}}}
\def\Mat#1{{\boldsymbol{#1}}}

\newcommand*{\Three}[1]{\parbox[c]{3cm}{\raggedright #1}}%
\newcommand*{\Two}[1]{\parbox[c]{2cm}{\raggedright #1}}%

\begin{document}

\title{Going Deeper into Action Recognition: A Survey}

\author{\name Samitha Herath \email samitha.herath@data61.csiro.au \\
       \addr College of Engineering and Computer Science\\
      Australian National University \& Data61, CSIRO\\
      Canberra, Australia
       \AND
       \name Mehrtash Harandi \email mehrtash.harandi@data61.csiro.au \\
       \addr College of Engineering and Computer Science\\
      Australian National University \& Data61, CSIRO\\
      Canberra, Australia
        \AND
       \name Fatih Porikli \email fatih.porikli@data61.csiro.au \\
       \addr College of Engineering and Computer Science\\
      Australian National University \& Data61, CSIRO\\
      Canberra, Australia}

\maketitle

\begin{abstract} 

Understanding human actions in visual data is tied to advances in complementary research areas including object recognition, human dynamics, domain adaptation and semantic segmentation. Over the last decade, human action analysis evolved from earlier schemes that are often limited to controlled environments to nowadays advanced solutions that can learn from millions of videos and apply to almost all daily activities. Given the broad range of applications from video surveillance to human-computer interaction, scientific milestones in action recognition are achieved more rapidly, eventually leading to the demise of what used to be good in a short time. This motivated us to provide a comprehensive review of the notable steps taken towards recognizing human actions. To this end, we start our discussion with the pioneering methods that use handcrafted representations, and then, navigate into the realm of deep learning based approaches. We aim to remain objective throughout this survey, touching upon encouraging improvements as well as inevitable fallbacks, in the hope of raising fresh questions and motivating new research directions for the reader.

\end{abstract}

%\begin{keyword}

%Human Action Recognition; Motion Recognition; Survey; Deep networks 
%\end{keyword}
%\linenumbers

\section*{Introduction}

Imagine the time when your smart environment and your robot assistant are capable of recognizing and understanding your actions at a level that they may actually help you in getting things done without your intervention. 

We may not be there yet, but our technological progress is geared evidently towards such a marvelous time. In this survey, we walk through existing research on action recognition in the hope of shedding some light on what is available now and what needs to be done in order to develop smart algorithms that are semantically aware of our actions.

\subsection*{\textbf{But first, what is an \emph{action}?}}

Human motions extend from the simplest movement of a limb to complex joint movement of a group of limbs and body. For instance, while the leg movement on a football kick is a simple motion, jumping for a head-shoot would be a collective movements of legs, arms, head, and whole body. Despite its intuitive and rather simple concept, the term \emph{action} seems  to be hard to define!
Below, we provide a few examples from the literature:

\begin{itemize}[noitemsep]

\item \cite{Moeslund_Surv06,Poppe_Surv10} define \textit{action primitives} as ``an atomic movement that can be described at the limb level''. Accordingly, the term \textit{action} defines a diverse range of movements, from ``simple and primitive ones'' to 
``cyclic body movements''. 
The term \textit{activity} is used to define ``a number of subsequent actions'', representing a complex movement. For instance, left leg forward is an action primitive of running. Jumping hurdles is an activity performed with the actions starting, running and jumping.

\item \cite{Turaga_Surv08} define \textit{action} as ``simple motion patterns usually executed by a single person and typically lasting for a very short duration (Order of tens of seconds).'' Their \textit{activity} refers to ``a complex sequence of actions performed by several humans who could be interacting with each other in a constrained manner.'' For example, actions are walking or swimming, activities are two persons shaking hands or a football team scoring a goal.

 ~\cite{Chaaraoui_HBA12} suggests a hierarchical breakdown of human motions in the context of human behavior analysis. The breakdown is based on the level of semantics and the temporal granularity, and considers the ``action'' in a level between the ``motion'' and the ``activity''. Actions are defined as primitive movements (\eg, sitting, walking) that can last up to several minutes.

\item \cite{WangGupta_ActTrans15} suggest that the true meaning of an \textit{action} lies in ``the change or transformation an action brings to the environment'', \eg, kicking a ball. 

\end{itemize}

In the Oxford Dictionary, \textit{action} is defined as ``the fact or process of doing something, typically to achieve an aim''. and \textit{activity} is ``a thing that a person or group does or has done''. We provide a consolidated definition that serves our purposes in this study the best.\\
\emph{``\textbf{Action} is the most elementary human
\footnote{ In this survey we are chiefly interested in human actions.Nevertheless, an action can be defined in a broader context by excluding the dependency on humans (\eg, actions performed by robots).}-surrounding interaction with a meaning.''}

The \emph{meaning} associated with this \emph{interaction} is called the \emph{category} of the \emph{action}. In general, human actions can take various physical forms. In our definition, the term \emph{interactions} can be understood as \emph{relative motions} with 
respect to the surrounding that may or may not cause a change. In some situations, one may need to associate ``surrounding'' to particular objects to derive a meaningful interpretation (\eg, brushing hair). This is aligned with the definition of~\cite{WangGupta_ActTrans15}, where an action is defined by the change it is brought to the environment.

\begin{figure}[h!]
\centering
\includegraphics[width=1\textwidth]{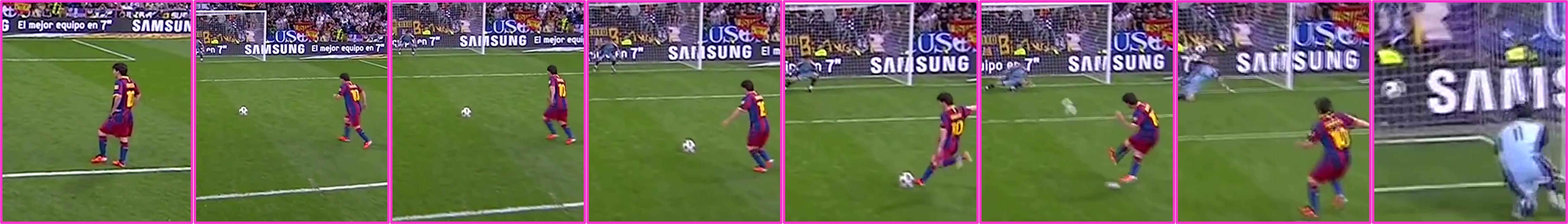}
\caption{Actions are ``meaningful interactions'' between humans and the environment.}
\label{fig:intro_action}
\end{figure}

As an example, consider the motion sequence in Fig.\ref{fig:intro_action}.
First, consider a primitive leg motion performed by the player on his run. Even though such movement is a relative motion with respect to the surrounding, we can barely attach a meaning to it. On the other hand, the collective motion of limbs, which results in running, has a meaning. Since this is the most elementary and meaningful motion, we consider it as an action, ``the running action''. Similarly, it is clear that the player's kick and the Jump of the goal-keeper are two distinct actions with labels ``kicking'' and ``jumping''. 

\subsection*{\textbf{Every survey paper has a taxonomy}}

True, but a generic taxonomy eludes us! Instead, we group solutions based on the fundamental understanding the reader will take at the end. 
We dedicate a separate section to deep learning based techniques where we discuss various architectures and training methods. At the same time, we arrange video representation based solutions, \ie, methods based on the handcrafted features, at the level of locality that their representations are constructed. We consider this dual nature of taxonomy is useful in highlighting essential components of the two categories.

\begin{figure}[h!]
\centering
\includegraphics[width=0.7\textwidth]{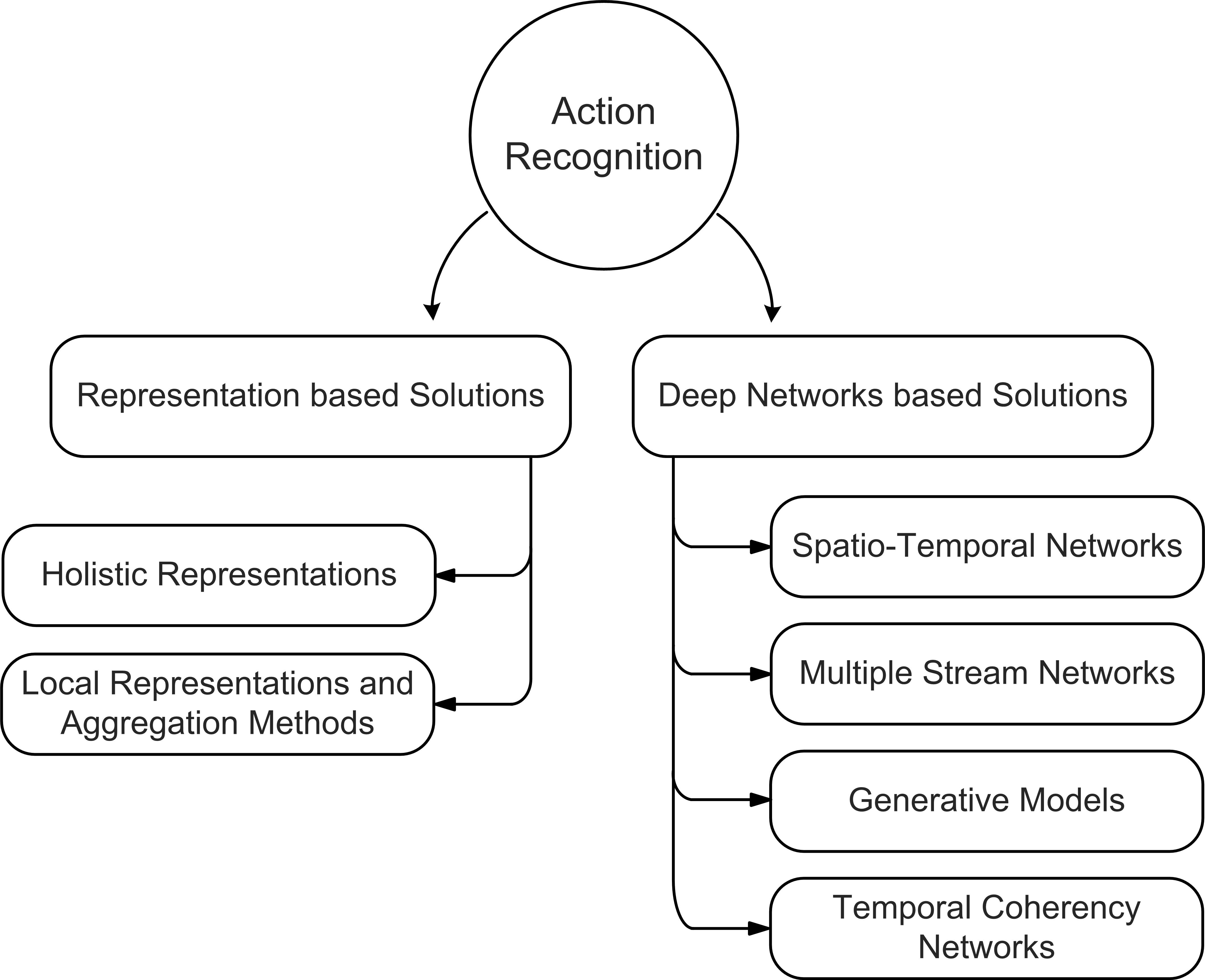}
\caption{A taxonomical breakdown we follow in this survey.}
\label{fig:taxonomy}
\end{figure}

To have a glance, the topics that will be covered in our study are shown in Fig.~\ref{fig:taxonomy}.
  
\subsection*{\textbf{Why should we learn more about action recognition?}}

Analyzing motions and actions has a long history and is attractive to various disciplines including psychology, biology 
and computer science (see Table.~\ref{tab:survey} for the list of surveys related to motion and action recognition in computer vision). 
One can trace the fascination about motion back to 500BC with Zeno's dichotomy paradox. 
From an engineering perspective, action recognition extends over a broad range of high-impact societal applications, from video surveillance to human-computer interaction, retail analytics, user interface design, learning for robotics, web-video search and retrieval, medical diagnosis, quality-of-life improvement for elderly care, and sports analytics. The long list of emerging technologies and applications 
(see for example~\cite{Ahad}) points to ``manually analyzing action and motion data is impossible''.

\begin{table}[h]
\small
\centering
\caption{Surveys on Motion and Action Analysis}
\label{tab:survey}
\begin{tabular}{ L{4.5cm} L{4.5cm} }
\hline
Survey & \multicolumn{1}{c}{Scope}  \\ \hline
~\cite{Moeslund_Surv06} & human motion capture and analysis  \\
~\cite{Yilmaz_Surv06} & object detection and tracking  \\
~\cite{Turaga_Surv08} & human actions, complex activities  \\
~\cite{Zhan_Surv08} & surveillance and crowd analysis \\
~\cite{Poppe_Surv10} & human action recognition \\
~\cite{Weinland_Surv11} & action recognition \\ 
~\cite{Aggarwal_Surv11} & motion analysis fundamentals  \\
~\cite{Chaaraoui_HBA12} &  human behavior analysis and understanding\\
~\cite{Metaxas_Surv13} & human gestures to group activities  \\
~\cite{Vishwakarma_Surv13} & activity recognition and monitoring \\\hline
\end{tabular}
\end{table}

%===================================================================================================================================

\section{Where to start from?} 
\label{sec:representation}

Let us begin by quoting a visionary thought from early eighties: \emph{``First, there must be a symbolic system for representing the shape information in the brain, and, secondly the brain must contain a set of processors capable of deriving this information from images''}~\cite{Marr_Criteria82}.
In the context of action recognition, a good representation must 
``be easy to compute'', ``provide description for a sufficiently large class of actions'', 
``reflect the similarity between two like actions'', and ``be robust to various variations (\eg, view-point, illumination)''.

Earliest works in action recognition make use of 3D models to describe actions. 
One notable example is the \emph{WALKER} hierarchical model introduced in~\citep{Hogg_3DMod83} to understand and interpret human actions.
Another example is the use of connected cylinders to model limb connections for pedestrian recognition~\cite{Rohr_Model94}. 

Generally speaking, constructing accurate 3D models from videos is difficult and expensive. Therefore, many solutions avoid 3D modeling and instead opt for representing actions at a holistic or local level\footnote{Action recognition from Motion Capture Systems (MoCap)
and RGBD data is an active line of research these days. Interested reader is referred to 
the work of ~\cite{Harandi_CVPR14,Vemulapalli_CVPR14,Koniusz_CP16,Rahmani_CVPR15} for the MoCap data and~\cite{Oreifej_CVPR13,Du_CVPR15,Rahmani_CVPR16,Liu_ECCV16} for the RGBD data.
}. 
Formally, we can define:

\begin{itemize}[noitemsep]

\item \textbf{Holistic representations.} Action recognition is based on the extraction of
a global representation of human body structure, shape and movements. 

\item \textbf{Local representations.} Action recognition is based on the extraction of local features.

\end{itemize}

\begin{figure}[h!]
\centering
\includegraphics[width=0.5\textwidth]{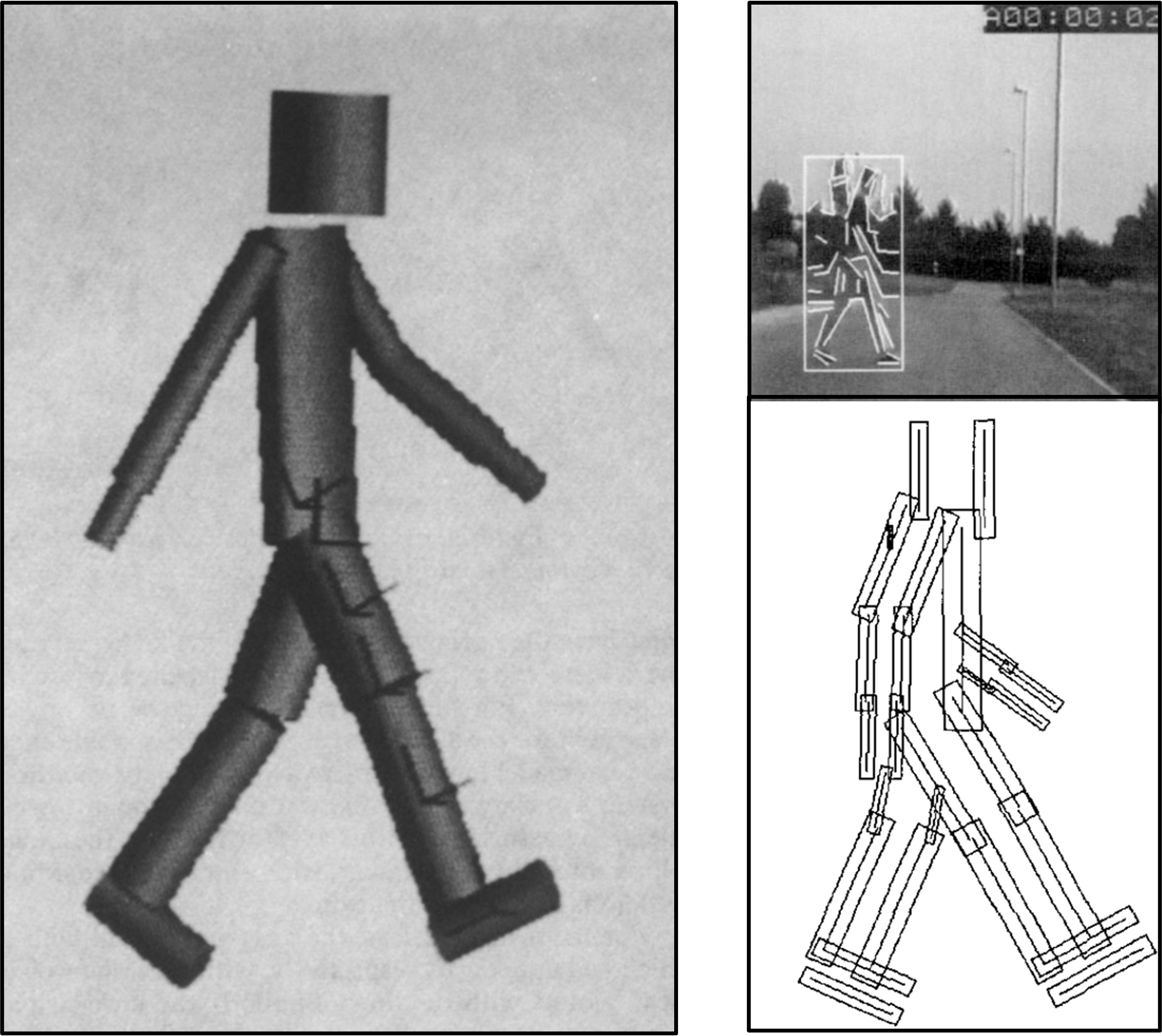}
\caption{Early approaches represent actions by 3D models. \textbf{Left:}~\cite{Hogg_3DMod83} introduce the \emph{WALKER} framework to represent walking action using 3D models. The walking pattern is modeled by a sequence of 3D structures. \textbf{Right:} ~\cite{Rohr_Model94} extended the WALKER framework for pedestrian recognition. The model uses connected cylinders and their evolution to identify pedestrians.}
\label{fig:3D_Models}
\end{figure}

\subsection{Holistic Representations}

We begin by describing the influential work of ~\cite{Bobik_MEI01}.
Motion Energy Image (MEI) and Motion History Image (MHI) are introduced in~\cite{Bobik_MEI01}. As the names suggest, the underlying idea is to encode the  motion-related information by a single image. The MEI template is a binary image describing where the motion happens 
and is defined by
\begin{align}
\Mat{E}_{\tau}(x,y,t) = \bigcup_{i=0}^{\tau-1}\Mat{D}(x,y,t-i)\;.\label{eqn:MEI}
\end{align}

\begin{figure}[h!]
\centering
\includegraphics[width=0.8\textwidth]{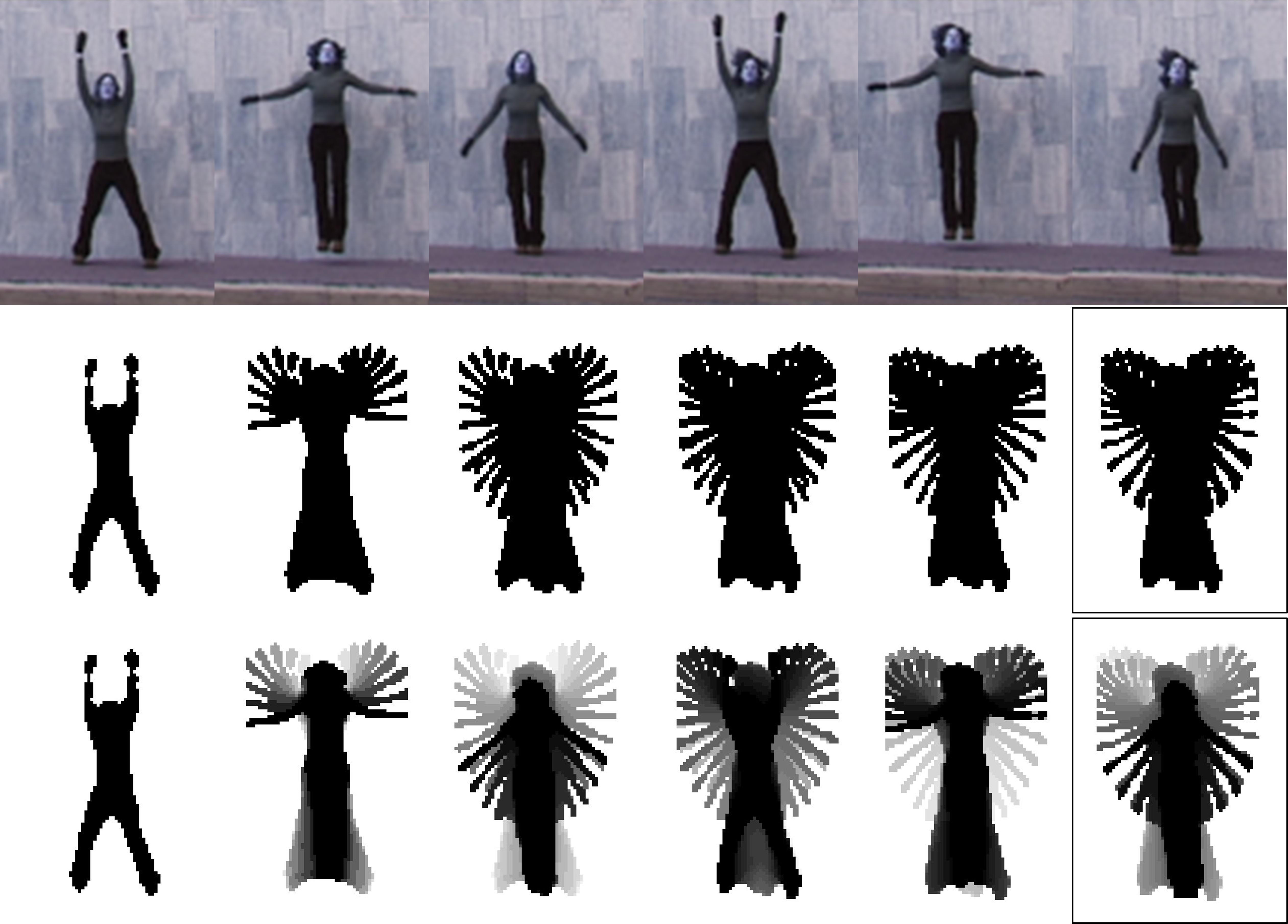}
\caption{\textbf{Top:} A jumping sequence. \textbf{Middle:} The MEI template~\cite{Bobik_MEI01}. \textbf{Bottom:} The MHI template~\cite{Bobik_MEI01}. The MEI captures where the motion happens while the MHI template shows how the motion image is moving. 
The templates at the end of the action, shown in the rightmost column are used for representations.}
\label{fig:MHI_MEI}
\end{figure}

Here, $D(x,y,t)$ is a binary image sequence representing the detected object pixels while $E_{\tau}$ denotes the formed MEI at a time $\tau$.
The MHI template shows how the motion image is moving. Each pixel in MHI is a function of the temporal history of
the motion at that point (\ie, higher intensities correspond to more recent movements) (see Fig.~\ref{fig:MHI_MEI} for an illustratation)

% MHI MHI

The MEI and MHI templates contain useful information about the context of videos. For example, the gradient of the MHI template is used to filter out the moving and cluttered background in~\cite{Tian_MHI12}. 
This is achieved by determining key motion regions in the MHI template using Harris interest point detector~\citep{Harris_88}, followed by identifying the moving/cluttered background as regions with inconsistent motions around the interest points.

\begin{figure}[h!]
\centering
\includegraphics[width=1\textwidth]{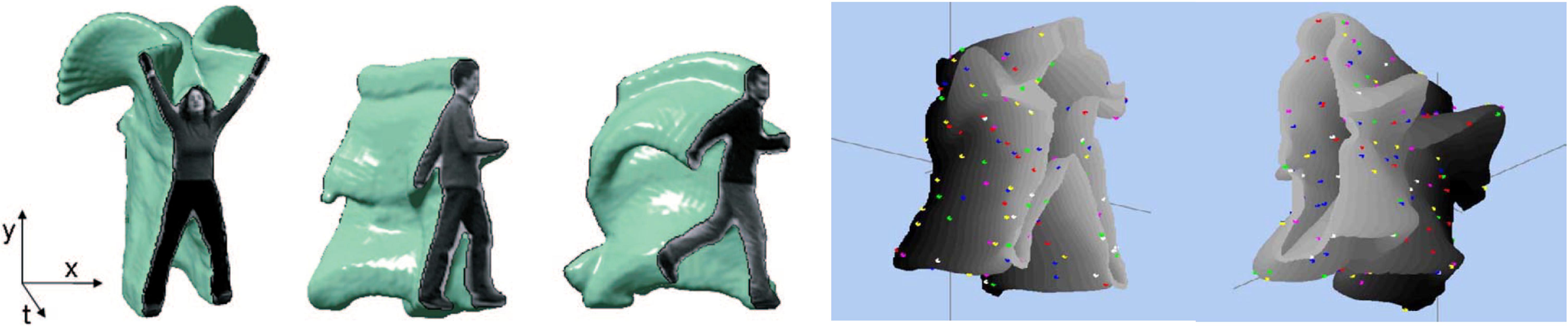}
\caption{\textbf{Left:} The spatiotemporal volumes used by~\cite{Blank_STShapes05} to describe the evolution of an action. The 3D representation is converted to a 2D map by computing the average time taken by a point to reach the boundary. \textbf{Right:} 
The spatiotemporal surfaces of~\cite{Yilmaz_STV05} for a tennis serve and a walking sequence. 
The surface geometry (\eg, peaks, valleys) is used to characterize the action.}
\label{fig:3D_Templates}
\end{figure}

The volumetric extension of MEI templates is introduced in~\cite{Blank_STShapes05}. The main idea is to represent an action by a 
3D shape induced from its silhouettes in the space-time (see Fig.~\ref{fig:3D_Templates}). For classification purposes, the resulting 3D surface is converted to a 2D map by computing the average time each point inside the surface requires to reach the boundary. 
A related study suggests to represent the MHI templates by spatiotemporal 
volumes~\citep{Weinland_3DMHV}, demonstrating extension to 3D volumes adds robustness to view point variations. 

~\cite{Yilmaz_STV05} propose to identify actions based on the differential properties of the \emph{Space-Time Volume} (STV). 
An STV is build by stacking the object contours along the time axis (see Fig.~\ref{fig:3D_Templates}). 
Changes in direction, speed and shape of an STV inherently characterize the underlying action. 
Action sketch is a set of properties extracted from the surface of an STV (\eg, Gaussian curvature) 
and is shown to be robust to view point changes. 

%3D Templates 

\subsubsection*{\textbf{A Statistically Correct Message}}

Holistic representations flooded the research in action recognition roughly between 1997 to 2007,
since such representations are more likely to preserve the spatial and temporal structure of actions. 
However, nowadays local and deep representations are favored~\citep{Wang_ImpTraj13,Peng_SFV14,Simonyan_Two14,KarpathyLarge14}.
Various reasons are attributed to this shift. For example,~\cite{Dollar_STIP05} claim that the holistic approaches are too rigid to capture possible variations of actions (\eg, view point, appearance, occlusions).~\cite{Matikainen_Traj09} believe that silhouette based representations are not capable of capturing fine details within the silhouette. As such, maybe it is time to change the gear and delve into local and deep solutions!

\section{Local Representation based Approaches}

Local representations for action recognition emerge as a result of the seminal work of ~\cite{Laptev_STIP05} on 
Space-Time Interest Points (STIPs). 
As in the case of images, local representations for action recognition follow the pipeline of 
\emph{interest point detection $\rightarrow$ local descriptor extraction $\rightarrow$ aggregation of local descriptors}.
Below, we review the key ideas and major developments for the aforementioned components separately.

\subsection{Interest Point Detection}

\begin{figure}[h!]
\centering
\includegraphics[width=0.7\textwidth]{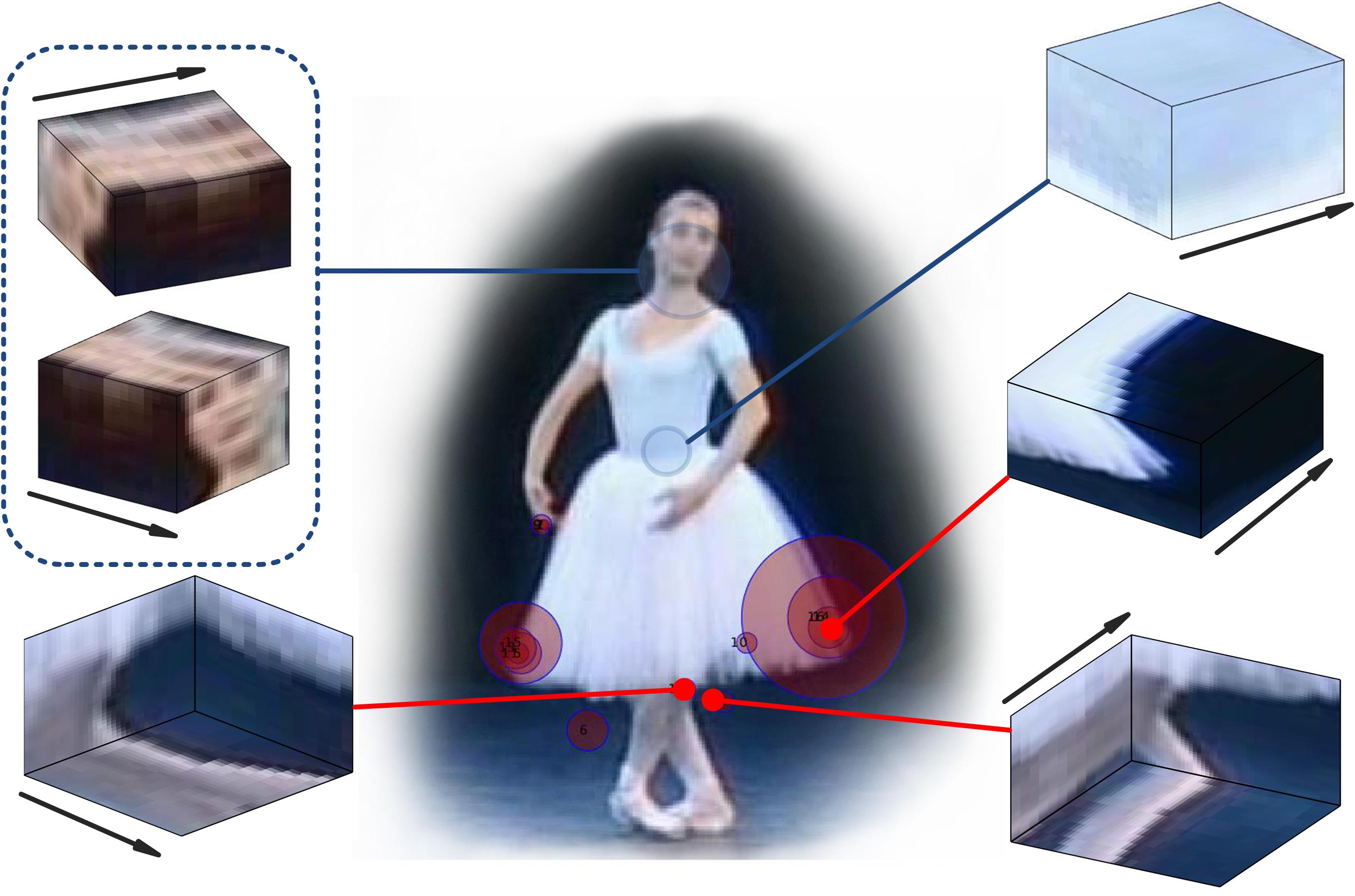}
\caption{Marked in red are the detected spatiotemporal interest points of ~\cite{Laptev_STIP05}. 
Spatial changes along the time axis (marked with an arrow) are noticeable. In this ballet video, the dancer keeps her head still throughout the video. Hence, despite having significant amount of spatial features, no spatiotemporal interest point is detected on the face. 
Similarly, in her waist no spatiotemporal interest point can be detected as a result of limited spatial variations.}
\label{fig:STIP}
\end{figure}

To build an STIP detector, ~\cite{Laptev_STIP05} extends the \emph{Harris} corner detector ~\citep{Harris_88} to \emph{3D-Harris}
detector. In 3D-Harris, in addition to rich spatial structures, temporal significance is required to fire the detector. 
The idea of the 2D Harris corner detector is to find spatial locations in an image with significant changes in two orthogonal directions.
The 3D-Harris detector identifies points with large spatial variations and non-constant motions.
An example of such requirements is shown in Fig.~\ref{fig:STIP}.

Another widely used 2D interest point detector, the \emph{Hessian} detector, is also extended to its 3D counterpart in~\cite{Willems_STIP08}. Unlike the 3D-Harris detector where gradients 
are  used towards detecting interest points, 3D-Hessian detector makes use of the second order derivatives for its decisions.

In certain domains, \eg, facial expressions, ~\cite{Dollar_STIP05} notice that true
spatiotemporal corners, as required by the 3D-Harris or 3D-Hessian detector, are quite rare, even if an interesting motion is 
occurring. While sparseness is desirable to an extent, STIPs that are too rare can lead to difficulties in action recognition. To overcome this limitation, in~\cite{Dollar_STIP05} it is proposed to disintegrate spatial filtering from the temporal one.
The resulting detector is shown to respond to any region with spatially distinguishing characteristics
undergoing a complex motion.

Unlike images, action clips are more likely to be obtained in uncontrolled environments. As such, 
care should be taken in processing videos  since the possibility of good features latching into irrelevant details is high. 
For example, a shaky camera can fire a series of irrelevant interest points. To address this issue,  ~\cite{Lui_VidWild09} 
suggest to prune irrelevant features using statistical properties of the detected interest points. 
Furthermore, spatiotemporal features obtained from background, known as \emph{static} features, 
especially the ones that are near motion regions are useful for action recognition~\citep{Lui_VidWild09}.
The relevance of static features for action recognition should not sound counter-intuitive.
This is because the background in certain types of videos (\eg, football) can provide useful contextual information
for action recognition. Moreover and from psychology we know that human beings are able to recognize many types of actions from still images without motion information.

\subsection{Local Descriptors}

Let us start with a simple definition, a 3D cuboid or simply a cuboid is a cube constructed from pixels around detected interest points. 
To obtain the local descriptor at an interest point, earlier works almost unanimously opt for 
cuboids~\citep{Dollar_STIP05,Laptev_STIP05}.  
In 2009,  separate studies by ~\cite{Messing_Traj09} and  ~\cite{Matikainen_Traj09} 
questioned the choice of fixed shaped cuboids for action recognition and introduced the notion of trajectories. 
Below, we first discuss various local descriptors widely used for action recognition, remembering that 
local descriptors can be employed with both cuboids and trajectories.
We then review trajectories and their improvements.

% STIP Harris3D Hessian 3D
\subsubsection*{\textbf{Edge and Motion Descriptors}}

~\cite{Klaser_HoG3D} suggest using the Histogram of Gradient Orientations as a motion descriptor.
% (see Fig.~\ref{fig:Hog3D} for details). 
While being inspired by the Histogram of Oriented Gradients (HoG)~\citep{Dalal_HoG05}, the descriptor itself is spanned to the spatiotemporal domain, hence named  the \emph{HoG3D} descriptor.

\begin{figure}[h!]
\centering
\includegraphics[width=1\textwidth]{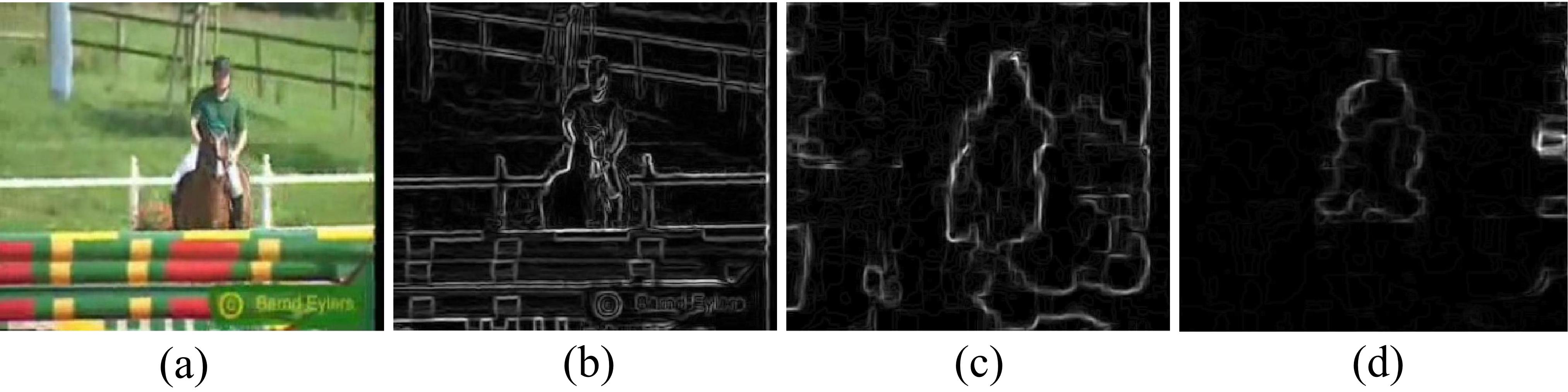}
\caption{The spatial gradients(b), horizontal(c) and vertical(d) motion boundary image for the horse riding action in (a). 
Unlike the spatial gradient which disregards motion information, the motion boundary images stress on the moving object boundaries.
Motion boundary images are obtained by computing the gradients of the optical flow fields.}
\label{fig:MBH}
\end{figure}

Optical flow fields encode the pixel level motions in a video clip. Exploiting this property,~\cite{Laptev_Realistic08} propose the Histogram of Optical Flow (HoF) over local regions as a spatiotemporal descriptor. 
A more robust extension of the HoF descriptor is the Motion Boundary Histogram(MBH) introduced in~\cite{Dalal_MBH06}. MBH is computed over the Motion Boundary fields, \ie, the spatial derivative of optical flow fields (see Fig.~\ref{fig:MBH} for an example).   
Though being rich, computing optical flow fields is computationally expensive. To overcome this difficulty, 
~\cite{Kantorov_MPEGFlow14} propose to make use of video decompression techniques. More specifically, instead of computing the optical flow fields for obtaining MBH or HoF descriptors, the authors use the motion fields in MPEG compression. 
This motion field, termed \emph{MPEG Flow}, can be obtained virtually free in the video decoding process.

\subsubsection*{\textbf{Pixel Pattern Descriptors}}
\begin{figure}[h!]

\centering
\includegraphics[width=0.8\textwidth]{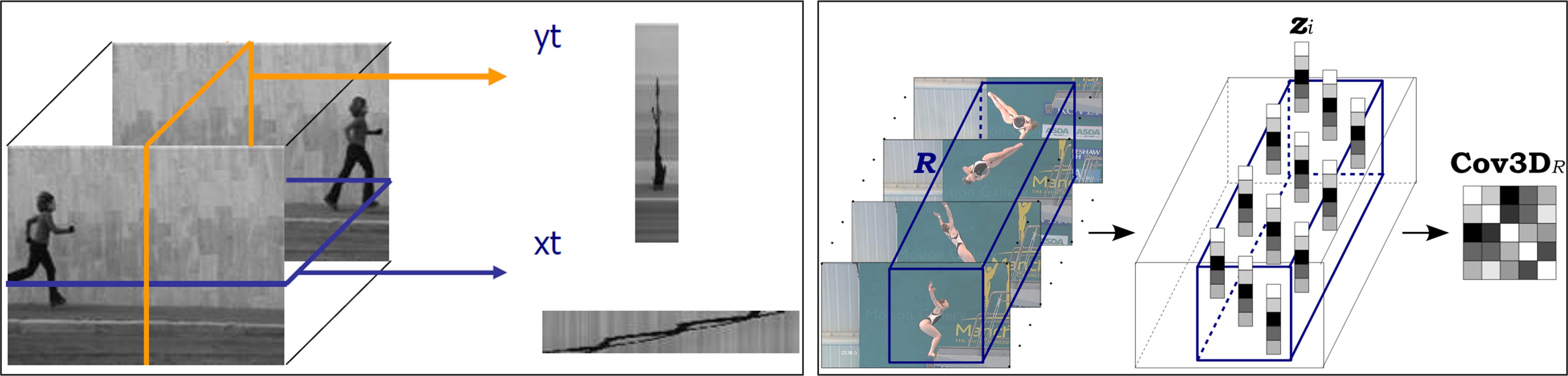}
\caption{\textbf{Left:} The LBP extraction planes of ~\cite{Kellokumpu_LBP08} for action recognition inspired by the LBP-TOP descriptor of ~\cite{Zhao_LBP07}. Here, the video stream is considered as a spatiotemporal volume and LBP descriptors are only extracted from the two orthogonal planes to the image plane \textbf{Right:} The spatiotemporal covariance descriptor of ~\cite{Sanin_Cov3D13}. 
Given a spatiotemporal window $R$, first a set of features $z_i\in \mathbb{R}^d$ is extracted from $R$ (dense or sparse). 
The spatiotemporal window is then described by the $d \times d$ covariance of the extracted features $z_i\in \mathbb{R}^d$.} 
\label{fig:Sparse_Traj_LBP_Cov3D}
\end{figure}

Local binary patterns (\emph{LBP}) are intensity-based 2D descriptors, successfully used in a diverse range of problems in vision including face recognition and texture analysis~\citep{Ojala_LBP02}. The LBP descriptor is computed by quantizing  the neighborhood of a pixel with respect to its  intensity. In~\cite{Zhao_LBP07}, various extensions of the 2D LBP descriptors to spatiotemporal domain are introduced. 
In the Volume LBP (VLBP), local volumes are encoded by the histogram of the binary patterns~\citep{Zhao_LBP07}.
Despite its simplicity, the number of distinct patterns produced by VLBP can become overwhelming for large neighborhoods. 
To alleviate this difficulty, in the Local Binary Pattern histograms from Three Orthogonal Planes (LBP-TOP),
the descriptor is obtained by concatenating local binary patterns on three orthogonal planes, 
namely $xy$, $xt$ and $yt$ planes (see Fig.~\ref{fig:Sparse_Traj_LBP_Cov3D} (left) for an illustration for the LPB variant of ~\cite{Kellokumpu_LBP08}). 
The idea of three orthogonal plane is extended by ~\cite{Norouznezhad_HoGNOP12} to nine symmetric planes.

{Describing image regions through second order statistics is proposed in~\cite{Tuzel_Cov06}. In particular, to describe a region $R$ in an image (see extensions to videos in~\cite{Sanin_Cov3D13}), first a set of features $\{\Vec{z}_i\}_{i = 1}^n,~\Vec{z}_i \in \mathbb{R}^d$ is extracted from $R$ (dense or sparse). Common choices here are low-level features (\eg, gradients, RGB intensities) or mid-level features (\eg, SIFT or HoG)~\citep{Carreira_TPAMI14}. The $d \times d$ covariance matrix of $\{\Vec{z}_i\}_{i = 1}^n$, usually referred to as Region Covariance Descriptor (RCD), is then used as the descriptor for $R$. Considering its natural Riemannian structure, RCDs are robust to scale and translation variations, and show resilency to noise~\citep{Tuzel_Pedestrian08} (see Fig.~\ref{fig:Sparse_Traj_LBP_Cov3D} (right) for an illustration).}

\subsection*{\textbf{From Cuboids to Trajectories}}

\begin{figure}
\centering
\includegraphics[width=0.5\textwidth]{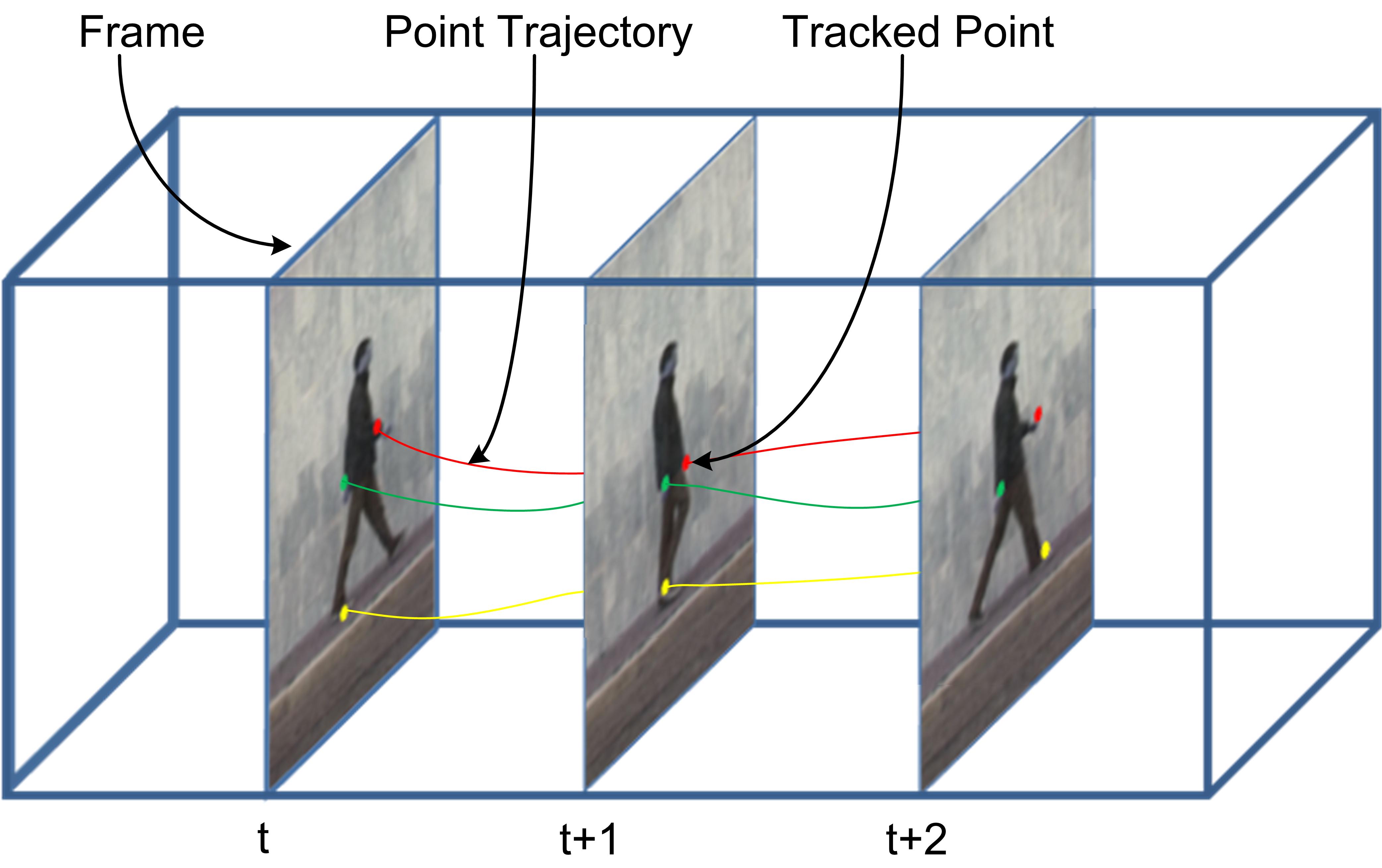}
\caption{Tracked point trajectories over frames.} 

\label{fig:Traj}
\end{figure}

An spatiotemporal interest point might not reside at the exact same spatial location within the temporal extends of a cuboid. Hence, features extracted from cuboids may not necessarily describing the interest point itself.
A trajectory is a properly tracked feature over time\footnote{In 1973  Johansson showed that human subjects could correctly
perceive ``point-light walkers'', a motion stimulus generated by a person walking in the dark, with points of
light attached to the its body. This study resembles the notion of trajectories.}, 
(see Fig.~\ref{fig:Traj}). Extracting local features from trajectories gains its popularity mostly from the work of ~\cite{Messing_Traj09} and ~\cite{Matikainen_Traj09}. Interestingly, both studies use a form of velocity of trajectories as local features.

Relative motions (\eg, differences in direction, magnitude and location) between trajectories can characterize certain action categories, especially, the categories that involve human/human interactions (\eg, hand-shaking) as shown by~\cite{Jiang_Traj12}.
Rectifying  trajectories using camera motions leads to improvements as shown in~\cite{Jiang_Traj12,Wang_ImpTraj13}. 
\cite{Jiang_Traj12} cluster trajectories to determine the dominant motion in a sequence. 
The dominant motion is assumed to be caused by the camera and is compensated from original trajectories by subtraction~\cite{Jiang_Traj12}
or through affine transformations~\cite{Jain_DCS13}. Nevertheless, both studies find the compensation may become misleading if 
a sizable portion of the video is covered by the actual action. 
The homography between consecutive frames is also used to estimate the camera motion\footnote{We note that Mikolajczyj 
and Uemura propose to make use of homography for compensating camera motions earlier~\cite{Mikolajczyj_Homogr08}.}~\cite{Wang_ImpTraj13}.

%Dense Trajectory Descriptors
\subsection*{\textbf{Sparse or Dense?}} 

In short, \emph{sparse is old, dense is new!} While early studies opt for sparse 
interest points, later, several  studies show the superiority of dense sampling 
in both image~\citep{Nowak_DenseSample06,FeiFei_Bayesian}
and video classification~\cite{Wang_Evaluation09}. A comprehensive comparison between various sparse  methods and dense sampling for several descriptors in action recognition can be found in~\cite{Wang_Evaluation09}.

\subsection{Aggregation}

Let $\mathbb{F} = \{\Vec{f}_i\}_{i=1}^n,~\Vec{f}_i \in \mathbb{R}^d$ be a set of local features extracted from a video.
For the purpose of action recognition, we need a mechanism to learn from such sets and eventually compare them.  
Learning algorithms such as Support Vector Machines (SVM) mostly accept fixed-size vectors and cannot work with sets of varying 
size (the number of local features varies per video). As such and in order to benefit from various learning techniques, we need a mechanism to aggregate sets of local features into discriminative and fixed-size descriptors. 
In doing so, machineries based on the concept of Bag-of-Visual Words (BoV)~\citep{Csurka_BOVW04} 
and \emph{Dictionary learning}~\citep{Olshausen_Sparse97,ELAD_SR_BOOK_2010} are the most natural choices.

\subsubsection{Aggregation with BoV}
In a nutshell, given a ``visual vocabulary'' or ``codebook'' $\mathbb{D} = \{\Vec{d}_j\}_{i=1}^k,~\Vec{d}_j \in \mathbb{R}^d$, the distribution of a given set of local descriptors $\mathbb{F} = \{\Vec{f}_i\}_{i=1}^n,~\Vec{f}_i \in \mathbb{R}^d$ on the codebook $\mathbb{D}$ is used as the descriptor.

In the BoV, the histogram of ``visual word'' occurrences is used as the descriptor. That is the frequency of seeing each 
visual word $\Vec{d}_j$ as the closest match to the local features $\Vec{f}_i$ determines the descriptor. 
The work of ~\cite{Dollar_STIP05} is among the first studies that resort to BoV for action recognition. 
In its original form, the temporal information is ignored by BoV. To ameliorate this shortcoming, 
~\cite{Laptev_Realistic08} propose the spatiotemporal grids. The main idea is to split a video 
into several sub-videos, aggregate the local descriptors of each sub-video to form the so-called ``channels'' and compare videos based on their channel descriptors. 
{An improvement inline with the concept of BoV is the hierarchical BoV~\citep{Kovashka_BoW10}. 
The base-level vocabulary is learned using HoG3D descriptors~\citep{Klaser_HoG3D}. 
Other levels of vocabulary are then constructed by aggregating their immediate lower level descriptors while spatiotemporal neighborhoods
are taken into account}.

More recently, aggregation through the Fisher Vector (FV) encoding~\citep{Wang_Traj11,Peng_SFV14,Oneta_FV13}
 becomes the method of choice. 
The FV encoding~\citep{Perronnin_FV07} is an aggregation method based on the principle of the Fisher Kernels~\citep{Jaakkola_FisherK98}
, which combines the benefits of generative and discriminative approaches to pattern classification. Briefly, the key differences between BoV and FV are 
\textbf{I)} BoV employs hard-assignment towards aggregation while FV benefits from soft-assignment, and
\textbf{II)} if the underlying model of feature generation is assumed to be a Gaussian Mixture Model, BoV only considers 
the zeroth-order information (occurrences) in aggregation while FV benefits from both first and second-order statistics. 
The FV encoding along trajectories delivered the state-of-the-art performances in several studies (see for example~\cite{Wang_Traj11,Wang_ImpTraj13}). Stacked FVs which can be understood as an extension of 
spatiotemporal grids of~\cite{Laptev_Realistic08} to FVs is introduced in~\cite{Peng_SFV14}. 
A detailed analysis of FVs in action recognition is presented in~\cite{Oneta_FV13}.

FVs are usually very high dimensional ~\citep{Jegou_VLAD10} and in certain applications redundant. A simplified version of 
FV, known as Vector of Locally Aggregated Descriptor (VLAD) ~\citep{Jegou_VLAD10,arandjelovic_VLAD13}, removes the second-order information from the descriptor. As a result, the dimensionality of VLAD descriptors is almost half of FVs. 
In~\cite{Jain_DCS13,Xing_VLAD15,Kantorov_MPEGFlow14,Sun_VLAD13}  VLAD descriptors obtained from spatiotemporal features are 
employed for action recognition. A comparison of speed and accuracy of FV against VLAD can be found in~\cite{Kantorov_MPEGFlow14,Wu_VLAD14}.

{
\subsubsection{Aggregation with Spatiotemporal Dictionary Learning and Sparse Coding}
}

{Sparse coding has become a popular choice in neuroscience, information theory, signal processing, and other related areas~\citep{Olshausen_Sparse97,CANDES_2006,Donoho_2006,ELAD_SR_BOOK_2010} in the last decade. By imposing sparsity, it is possible to represent natural signals such as images using only a few non-zero coefficients, \ie as a linear decomposition using a using a few atoms of a suitable dictionary. In computer vision, sparse image representations were originally introduced for modeling spatial receptive fields of cells in the human visual system by~\citep{Olshausen_Sparse97}. Following studies have been shown to deliver notable results for various visual inference tasks, such as face recognition~\citep{Wright:PAMI:2009}, subspace clustering~\citep{Elhamifar_2013_PAMI} and image restoration~\citep{Mairal_TIP_2008} to name a few.}

{For action recognition, \cite{Zhu_ACCV10} use the principals of sparse coding to aggregate local spatiotemporal features. Using a learned dictionary, they encode HoG3D descriptors obtained from uniformly distributed spatiotemporal cuboids. They obtain the video descriptor by performing max-pooling on the sparse codes. Moreover, to learn the dictionary, they suggest transfer learning from unlabeled video data.

\cite{Guha_PAMI12} study various forms of dictionaries for the task of action recognition. In the simplest form, a common dictionary across all action classes is learned. This common dictionary is shown to be limited in its representative power when new action classes are introduced. To alleviate this limitation,  the use of class specific dictionaries is suggested. 

To extract spatiotemporal feature, ~\cite{Somasundaram_CVIU14} suggest salient spatiotemporal regions. The main idea, inspired by the principals of information theory, states that the saliency of a spatiotemporal region is captured through its structural complexity\footnote{We note that a similar concept is used in~\cite{Laptev_STIP05} to identify spatiotemporal interest points.}. Through the use of a dictionary, the structural complexity of a patch is approximated by the concept of \emph{minimum description length}(MDL) ~\citep{RISSANEN_Auto78}. Intuitively, the number of bits required to represent it decreases as regularity in the data increases.

Inspired by the object bank method~\citep{Jia_ObjectBank10}, ~\cite{Sadanand_ActionBank12} propose the ``action bank'', where actions are described by a large set of detectors acting as the dictionary of a high-dimensional ``action-space''.  We point out that \emph{the action bank} itself is a high-level \emph{dictionary}.  A relevant idea is presented by ~\cite{Shao_Pyramids14} where the Laplacian of 3D Gaussian filters is used to construct the action space . Both previous methods exploit the pyramid structure to enhance robustness across spatial and temporal domains.}

\subsubsection{Aggregation via Temporal Coherence}

We conclude this part by describing studies that explicitly incorporate temporal information in aggregating spatiotemporal information for video descriptors. ~\cite{Fernando_VidDarwin15} propose to represent a video by a ranking machine. That is, given the frame descriptors, a hyperplane that ranks the frames according to their temporal order is used to represent the video. ~\cite{Gaidon_Actom11} propose the concept of atomic-actions or \emph{actoms} which can be understood as a temporally structured extension of the BoV. An actom\footnote{A relevant idea to the concept of actoms is proposed by Niebles \etal earlier~\cite{Niebles_TempStruct10}.} is the building block of an action and has a variable temporal extend. Histograms of visual words, similar to the traditional BoV approach, is used to describe an actom. Additionally, features that are located in the middle of an actom receive higher weights towards generating the histogram while the contribution of features away from the center is attenuated.

{
Other notable lines of research that exploit temporal coherence are Hidden Markov Models (HMMs) ~\citep{Rabiner_HMM89} and Conditional Random Fields (CRFs) ~\citep{Lafferty_CRF01} that are typically used for sequence modeling applications such as Natural Language Processing ~\citep{McCallum_NLP03} and Speech Processing ~\citep{Gales_Sigpro07}. An action could be interpreted as a sequence of appearance transitions ~\citep{Tang_CVPR12, Hongeng_ICCV03}. Hence, it is straightforward to model them using a state transition structure such as an HMM.

~\cite{Hongeng_ICCV03} uses HMMs in modeling events in videos. However, they break free from the inherent first order dependency between states of HMMs by using the mentioned \emph{semi-HMM}s. The work in ~\cite{Tang_CVPR12} employs the max-margin framework for modeling the latent temporal structure of a video. Furthermore, it adopts a \emph{Variable-duration HMM} which additionally associates a latent duration variable along with other latent state variables. In ~\cite{Sun_ICCV13}, the video is treated as a sequence of short clips corresponding to observations of HMM latent states. The objective of the model is to capture the complex activities by considering actions as such short clips.

In comparison, the discriminatively trained CRF models contain a distinct advantage over their generatively trained counterparts, \emph{i.e.} HMMs. Unlike the first order dependency of HMMs, CRFs are conditioned on the entire sequence. The work of ~\cite{Quattoni_TPAMI07} embeds latent variables into the CRF modeling and introduces the Hidden CRF (HCRF) for spatiotemporal recognition. In ~\cite{Wang_CRF11} applies \emph{max-margin} learning in modeling. A hierarchical CRF modeling approach based on the temporal granularity is presented in ~\cite{Song_CVPR13}.

The work in~\cite{Li_CVPR16} makes the observation that videos usually contain distinct scenes where dynamics are coherent only within them.  Therefore, it proposes hierarchical temporal models that are learned on three levels of temporal granularity where the levels are trained using CNN features (we discuss in the next section), linear dynamical systems, and VLAD codes. To capture the nonstationary evolution of dynamics in a video, a hierarchical encoding method is described in \cite{Su_ECCV16} where they suggest the \emph{Hierarchical Dynamic Parsing and Encoding} pipeline with two or more temporal encoding layers. 

We point the reader to the work of \cite{Kovashka_BoW10,Fernando_HRPCVPR16,Shao_Pyramids14} for similar ideas on hierarchical video analysis.
} 

\vspace{8mm}

%===================================================================================================================================

\section{Deep Architectures for Action Recognition}
\label{sec:deep_nets}

We are witnessing a significant advancement in countless tasks thanks to the deep and data driven architectures. Deep neural networks such as Convolutional Neural Networks (CNN)~\citep{Lecun_Conv98} have become the method of choice in learning image contents~\citep{krizhevsky_imagenet_2012,Chatfield_VGG14,Sutskever_NIPS14,Szegedy_GoogleNet15}. Generally speaking, the problem of learning is to determine a complicated decision function from the available data. In deep architectures this is achieved by composing multiple level of nonlinear operations. Searching the parameter space of deep architectures is not an easy job given the non-convexity of the decision surface. Learning algorithms based on the gradient descent approach along the computational power of new hardware have been shown to be successful when large amount of annotated data is available~\citep{Limin_GoodPrac15,Srivastava_VeryDN15,Kaiming_Residual15}.

Our intention in this section is to discuss deep models that have been used (or can potentially be used) to address the problem of learning actions from videos. From a taxonomical point of view, we can identify four categories of architectures applied to action recognition, namely,
\begin{itemize}[noitemsep]
\item Spatiotemporal networks
\item Multiple stream networks
\item Deep generative networks
\item Temporal coherency networks
\end{itemize}
Below, we discuss each category in detail and provide pointers to open questions and possible improvements.

\subsection{Spatiotemporal Networks}
\label{subsec:deep_spatio}

\begin{figure}[b!]
\centering
\includegraphics[width=0.8\textwidth]{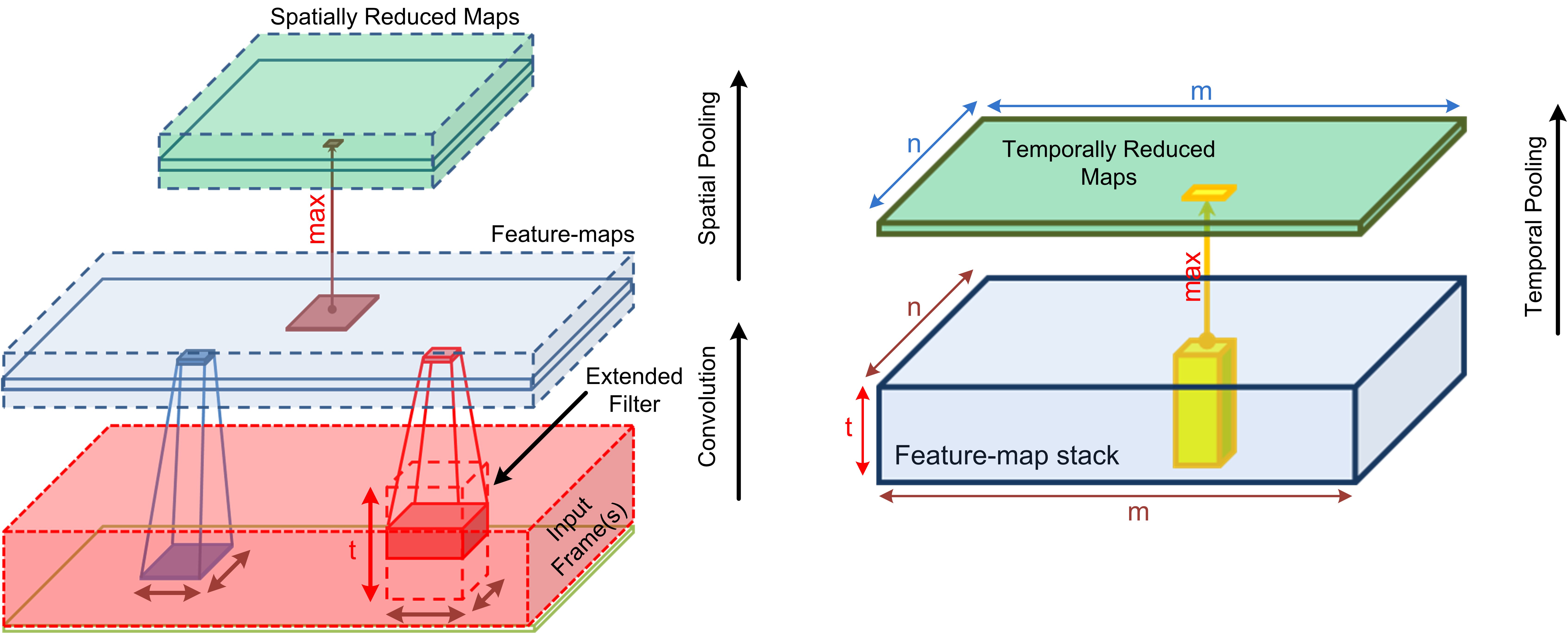}
\caption{Spatiotemporal operations: 2D convolution (blue), 3D convolution on frame stacks (red) as in ~\cite{Ji_3dConv10}, conventional spatial max-pooling (brown), and temporal max-pooling (yellow) as in ~\cite{Yue-Hei_Beyond15}.}
\label{fig:3DConvNet}
\end{figure}

The convolutional architecture effectively utilizes the image structure in reducing the search space of the network by ``pooling'' and ``weight-sharing'' (see Fig.~\ref{fig:3DConvNet} (left) for a conceptual diagram). Pooling and weight-sharing also contribute to achieving robustness across scale and spatial variations. Analyzing filters learned by CNN architectures suggests that the very first layers learn low level features (\eg, \emph{Gabor}-like filters) while top layers learn high level semantics~\citep{Zeiler_Visual14}. This further extends the use of convolutional networks as generic feature extractors. 

A direct approach to action recognition using deep networks is to arm the convolutional operation with temporal information. To achieve this, \emph{3D convolutional} networks are introduced in~\cite{Ji_3dConv10}. A 3D convolution network, as the name suggests, uses 3D kernels (filters extended along the time axis) to extract  features from both spatial and temporal dimensions, hence is expected to capture spatiotemporal information and motions encoded in adjacent frames (see Fig.~\ref{fig:3DConvNet} for a conceptual diagram). In practice, it is important to provide the network with supplementary information (\eg, optical flow) to facilitate training. Empirically,~\cite{Ji_3dConv10} show that the 3D convolutional networks outperform the 2D frame based counterparts with a noticeable margin.

Generally speaking, the 3D convolutional networks have a very rigid temporal structure. The network accepts a predefined number of frames as the input (for example in~\cite{Ji_3dConv10} the input consists of only $7$ frames). While having fixed spatial dimension is somehow defensible (spatial pooling tends to provide robustness across scales), it is unclear why a similar assumption should be made across the temporal domain. Even less clear is the right choice of the temporal span as macro motions in different actions have different speeds and hence different spans. 
 
\begin{figure}[t!]
\centering
\includegraphics[width=0.85\textwidth]{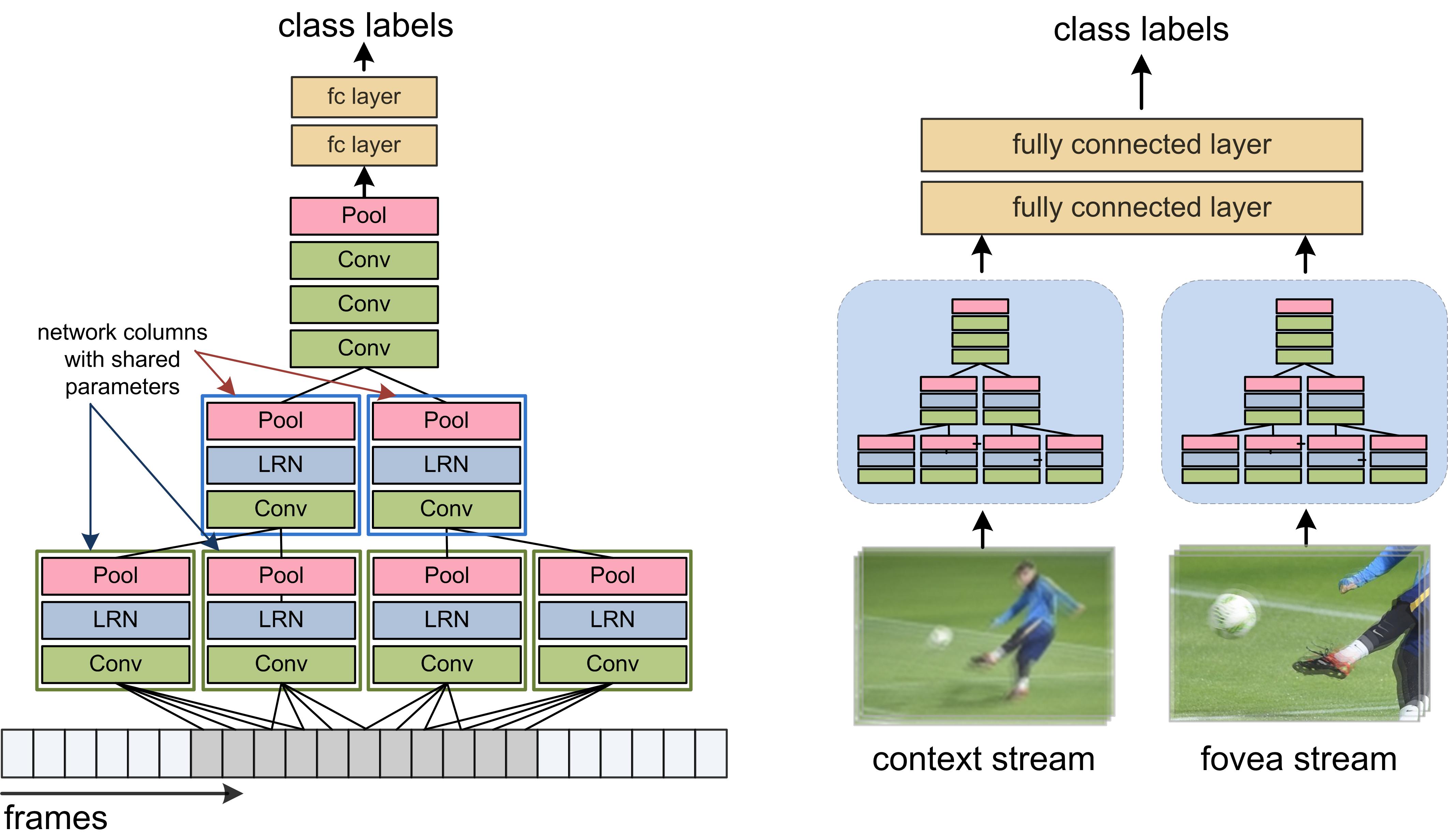}
\caption{The \emph{foveated} architecture of ~\cite{KarpathyLarge14}. Denoted in green, red and blue are respectively normalization, spatial-pooling and convolutional layers. }
\label{fig:Karapathy}
\end{figure}

To answer how temporal information should be fed into convolutional networks, various fusion schemes are investigated. \cite{Yue-Hei_Beyond15} explored temporal pooling and concluded that max pooling in the temporal domain is preferable. \cite{KarpathyLarge14} proposed the concept of \emph{slow fusion} to increase the temporal awareness of a convolutional network. In slow fusion, a convolutional network accepts several, yet consecutive, parts of a video and processes them through the very same set of layers to produce responses across temporal domain. These responses are then processed by fully connected layers to produce the video descriptor (see Fig.~\ref{fig:Karapathy} for details).

Other forms of fusion include \emph{early fusion} (\eg, the 3D convolutional network~\cite{Ji_3dConv10}) where the network is fed with a set of adjacent frames and \emph{late fusion} where frame-wise features are fused at the last layer~\citep{KarpathyLarge14}. ~\cite{KarpathyLarge14} also show that a multi-resolutional approach using two separate networks, not only boosts the accuracy but also reduces the number of parameters to be learned. This is due to the fact that each leg of the network (\ie, fovea and context streams in Fig.~\ref{fig:Karapathy}) accepts smaller inputs. We note that the fovea stream receives the central region of a frame to take advantage of
the camera bias that exists in many videos since the object of interest often occupies the central region.

{Similar to the use of \emph{VGG}~\citep{Chatfield_VGG14} and \emph{Decaf}~\citep{Donahue_ICML2014} networks as generic descriptors  for images,~\cite{Tran_C3D15} attempt to  find generic video descriptors based on a 3D convolutional network. The feature extraction network is trained on Sports-1M ~\citep{KarpathyLarge14} dataset. Empirically, the authors show that a network with $3 \times 3 \times 3$ homogeneous filters (constant depth at every layer) performs better than varying the temporal depth on filters. Flexibility on the temporal extent is obtained with the inclusion of 3D pooling layers. A generic descriptor named \emph{C3D}, is then obtained by averaging the outputs of the first fully connected layer of the C3D network.

\cite{varol_LTC16} explore the effect of performing 3D convolutions over longer temporal durations at the input layer. Improvements are observed by extending the temporal depth of the input as well as combining the decision of networks with different temporal awareness at the input.}

Extending spatial filters to 3D ones, though being mainstreamed, inevitably increases the number of parameters of the network. 
In ameliorating the downside effect of 3D filters,~\cite{Sun_ICCV15} suggest factorizing a 3D filter into a combination of a 2D and 1D filters.
{With this reduction of parameters, they obtain comparable performance to ~\cite{Simonyan_Two14} without any knowledge transfer between several video datasets while training (see Section $3.2$ for details on the knowledge transfer of ~\cite{Simonyan_Two14}).}

\begin{figure}[t!]
\centering
\includegraphics[width=0.6\textwidth]{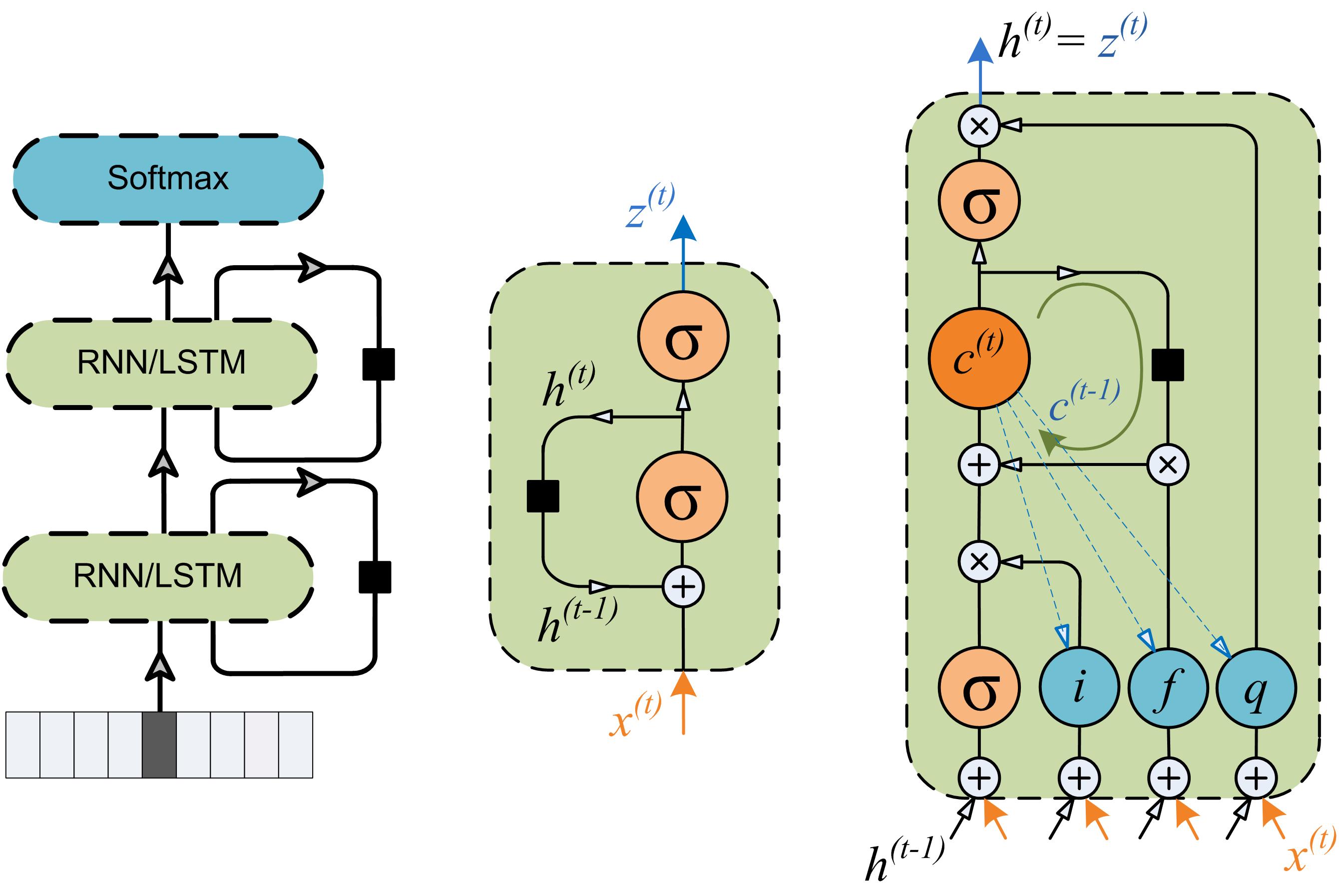}
\caption{\textbf{Left:} The recurrent structure of a 2-layer RNN/LSTM network. \textbf{Center:} The RNN cell structure that replicates a linear dynamical system. \textbf{Right:} The LSTM cell that includes additional gate controls. Time delay is indicated with the black square.}
\label{fig:RNN_LSTM}
\end{figure}

To exploit the temporal information, some studies resort to the use of recurrent structures. The works of \cite{Baccouche_Seq11} and \cite{Donahue_LRCN15} tackle the problem of action recognition through a cascade of convolutional networks and a class of Recurrent Neural Networks (RNN)~\citep{Robinson_RNN88} known as Long-Short Term Memory (LSTM) \citep{Hochreiter_LSTM97} networks. As the word \emph{recurrent} suggests, an RNN (see Fig.~\ref{fig:RNN_LSTM}) models the dynamics using a feedback loop. The typical form of an RNN block accepts an external signal $\Vec{x}^{(t)} \in \mathbb{R}^d$ and produces an output $\Vec{z}^{(t)} \in \mathbb{R}^m$ based on its hidden-state $\Vec{h}^{(t)} \in \mathbb{R}^r$ by
\begin{align}
\Vec{h}^{(t)} &= \sigma(\Mat{W}_{x}\Vec{x}^{(t)} + \Mat{W}_{h}\Vec{h}^{(t-1)})\;, \label{eqn:RNN_state}\\
\Vec{z}^{(t)} &= \sigma(\Mat{W}_{z}\Vec{h}^{(t)})\;.\label{eqn:RNN_output}
\end{align}
Here, $\Mat{W}_x \in \mathbb{R}^{ r \times d }$ , $\Mat{W}_h \in \mathbb{R}^{ r \times r }$ and $\Mat{W}_z \in \mathbb{R}^{ m \times r }$ . Obviously, an RNN is a realization of the Linear Dynamical Systems (LDS)~\citep{Huang_CVPR16}
and hence rich enough to model video sequences.

Generally speaking, training an RNN is not easy due to the issue of 
vanishing (or exploding) gradient ~\citep{Bengio_Vanishing94}.
For the sake of discussion, assume the recursive expression of an RNN cell has the form $h^{(t)} = w_h h^{(t-1)}$ with $x,h,z \in \mathbb{R}$. This recursive form can be unfolded as 
$h^{(t)} = w_h h^{(t-1)} = w_h w_h h^{(t-2)} = \cdots = w_h^t h^{(0)}$. As such, the network either learns short term dependencies (if $w_h < 1$) or very long dependencies (if $w_h > 1$) which is not desirable~\citep{Bengio_Vanishing94}. LSTM cells (shown in Fig.\ref{fig:RNN_LSTM}) solves this issue by constraining the states and outputs of the RNN cell through control gates. 

\begin{figure}[t!]
  \centering
  \includegraphics[width=0.75\textwidth]{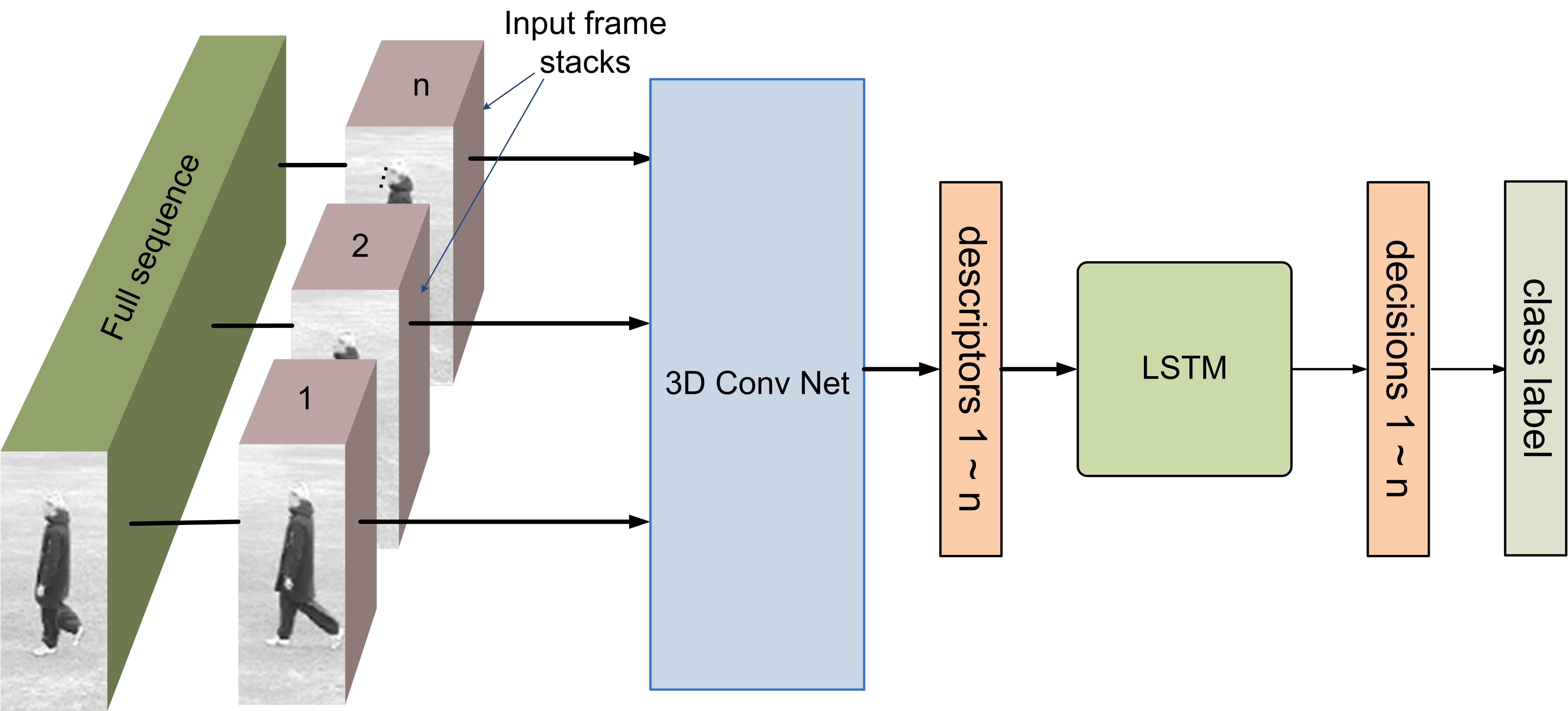}
  \caption{The network structure of \citep{Baccouche_Seq11}.}
  \label{fig:Baccouche_Network}
\end{figure}

To classify actions,~\cite{Baccouche_Seq11} suggest to feed an LSTM network with features extracted from a 3D convolutional network. The two networks, \ie, 3D convolutional network and the LSTM network are trained separately. That is, first the 3D convolutional network is trained using annotated action data. Once the 3D convolutional network is obtained, the convolutional features are used to train the LSTM network (see Fig.~\ref{fig:Baccouche_Network} for the network structure).

Another architecture based on LSTM is proposed by~\cite{Donahue_LRCN15} to exploit end-to-end training over the composite network as shown in Fig.~\ref{fig:LRCN_TwoStream}. The resulting structure named Long-term Recurrent Convolutional Network (LRCN) has been shown to be successful not only in recognizing actions but also in captioning images and videos. With the end-to-end learning and CNN-LSTM convolution, the spatiotemporal receptive filter parameters are computed in a data driven fashion.

\subsection{Multiple Stream Networks}  
\label{subsec:deep_multiple_streams}

In visual perception, the \emph{Ventral Stream} of our visual cortex processes object attributes such as appearance, color and identity. The  motion of an object and its location is  handled separately through the \emph{Dosaral Stream} ~\cite{goodale_1_2003}. A class of deep neural networks is devised to separate appearance based information from  motion related ones for action recognition~\cite{Simonyan_Two14}.

\begin{figure}[b!]
  \centering
  \includegraphics[width=1\textwidth]{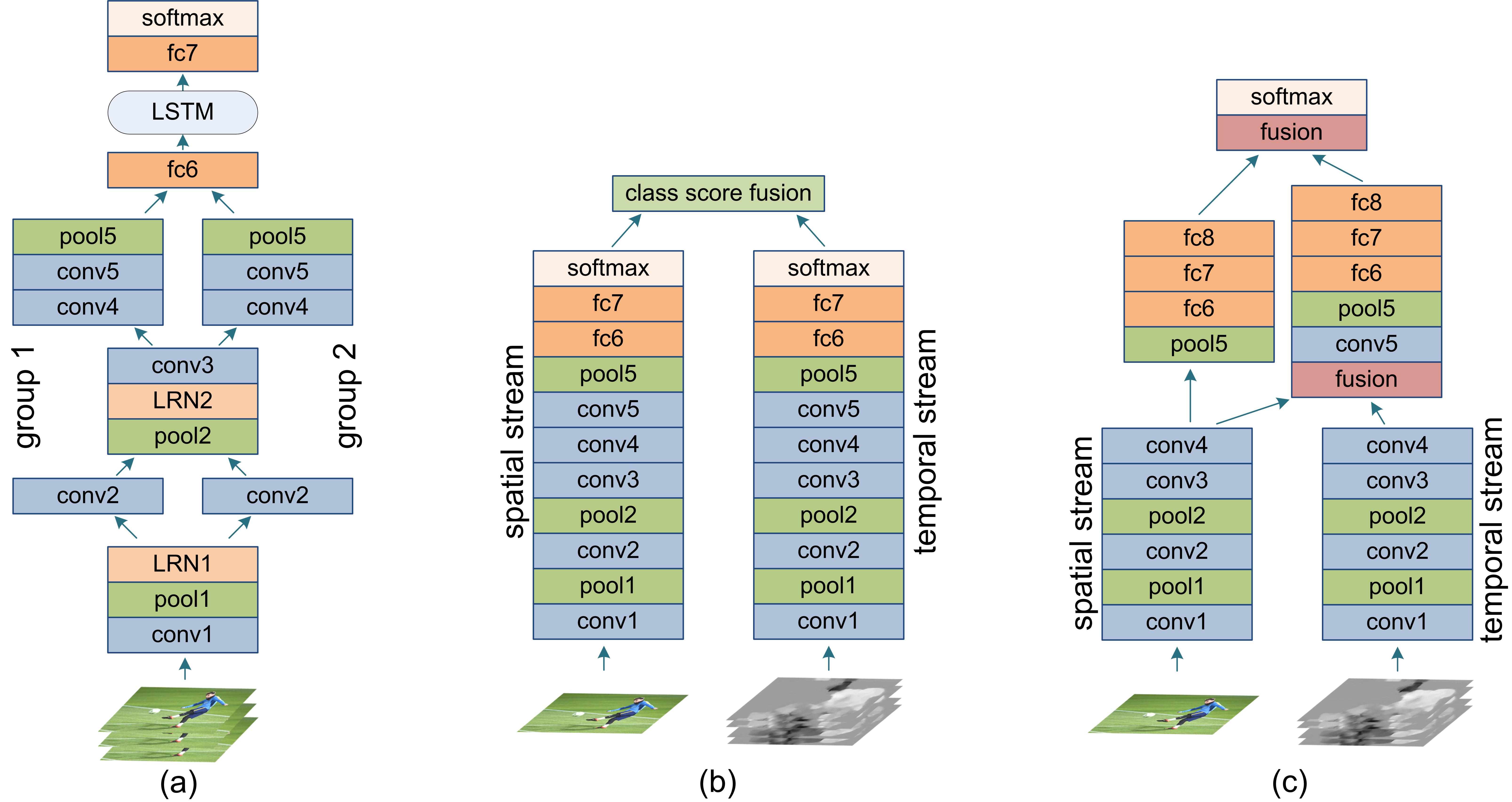}
  \caption{(a)~LRCN network structure of~\citep{Donahue_LRCN15}. {A \emph{group} is a  set of convolutional filters operating only on a particular set of feature maps from the previous layer. For clarity, we denote each group by a separate convolutional blocks.} 
(b) The two-stream network by ~\cite{Simonyan_Two14} with RGB and stacked optical-flow frames as inputs. (c) An example of a two stream fusion network of ~\cite{Feichtenhofer_CVPR16}.}
  \label{fig:LRCN_TwoStream}
\end{figure}

\cite{Simonyan_Two14} introduced one of the first multiple-stream deep convolutional networks where two parallel networks are used
for action recognition (see Fig.~\ref{fig:LRCN_TwoStream}). The so called spatial stream network accepts raw video frames while the temporal stream network gets optical flow fields as input. The following observations are made in~\citep{Simonyan_Two14}:

\begin{itemize}[noitemsep]

\item \textbf{Pretraining for the spatial stream network.} Training the spatial stream network from scratch is not the best practice. Empirically, fine-tuning a pretrained network on the ILSVRC-2012 image dataset ~\citep{ILSVRC15} leads to higher accuracy.

\item \textbf{Early fusion for the temporal stream network.} Stacking optical flow fields at the input of the temporal stream network (\ie, early fusion) is beneficial.

\item \textbf{Multi-task learning for the temporal stream network.} The temporal stream network needs to be trained purely from the available video data. This was observed to be challenging for small and medium-size datasets in very deep networks. To circumvent this difficulty, the temporal stream network is modified to have more than one classification layer. Each classification layer operates on a specific dataset (\eg, one operates on the HMDB-51 and one on the UCF-101 dataset) and responds only to the videos coming from the respective dataset. This architecture is a realization of the multi-task learning, aiming to learn a representation, which is not only applicable  to the task in question, but also to other tasks.
\end{itemize}
{The two streams are fused together using the softmax scores.  The work of~\cite{Feichtenhofer_CVPR16} shows that a fusion at an intermediate layer not only improves the performance but also reduces the number of parameters significantly (see Fig.~\ref{fig:LRCN_TwoStream} for an illustration). It demonstrates that the best accuracy is attained when the fusion is performed after the last convolutional layer. Interestingly, having the fusion right after the convolutional layers will remove the requirement of costly fully connected layers in both streams. Compared to the original network~\citep{Simonyan_Two14}, the fused network performs equally well with using only half the parameters.

Extensions of the two stream network include the work of~\cite{Wang_TDD15} where dense trajectories~\citep{Wang_ImpTraj13} traced over convolutional feature maps of the two-stream network are aggregated using the Fisher vector, and ~\cite{Wu_Fusing15} where a third stream using audio signal is added to the network. 

The optical flow frames are the only motion related information used in two stream networks. This would raise the question whether two stream networks can capture subtle but long-term motion dynamics (such motions cannot be modeled by optical flow). The improvements brought by effective combination of deep architectures and handcrafted solutions hint that certain details in actions are still out-of-reach in deep solutions~\cite{Feichtenhofer_CVPR16}.}

\subsection{Deep Generative Models}
\label{subsec:deep_generative}

The potential reward of devising deep models that require little or no supervision is beyond imagination, given the vast and ever increasing videos available on the Web. 
A good generative model is the one that can learn the underlying distribution of data accurately. Generative models for sequence analysis ~\citep{Sutskever_NIPS14,Srivastava_LSTM15} are mainly used to predict the future of a sequence. That is, given a sequence $\{\Vec{x}_1,\Vec{x}_2,\cdots,\Vec{x}_t\}$, one may deem to learn a model to  predict its future (\eg, the next instance $\Vec{x}_{t+1}$). This task is different from methods discussed in Section \textsection~\ref{subsec:deep_spatio} in nature as it does not require labels for training. However, accurate predictions are achieved if contents and dynamics (\eg, motion primitives) of the sequence can be captured by the model to a good extent. Deep-generative architectures~\cite{Vincent_Auto08,Goodfellow_Advers14,Hochreiter_LSTM97} aim this goal, \ie, learning from temporal data in an unsupervised matter. In video analysis where annotating data is costly, unsupervised techniques are preferred over supervised ones. 
Envisaging possible potentials, in this part we review notable examples of deep generative architectures without confiding ourselves to studies that have been directly applied to action recognition.  

\subsubsection{Dynencoder}

Inspired by LDS modeling~\cite{Doretto_DynTex03},~\cite{Yan_Dynencoder14} introduce \emph{Dynencoder}, a class of deep auto-encoders, to capture video dynamics. In its most basic form, a Dynencoder constitutes of three layers. The first layer maps the input $\Vec{x}_t$ to the hidden states $\Vec{h}_t$. The second layer is a prediction layer that predicts the next hidden states, $\tilde{\Vec{h}}_{t + 1}$, using current ones (\ie, $\Vec{h}_t$). The final layer is a mapping from the predicted hidden states $\tilde{\Vec{h}}_{t + 1}$ to generating estimated input frames $\tilde{\Vec{x}}_{t + 1}$. To reduce the training complexity, the parameters of the network are learned in two stages. In the pretraining stage, each layer is trained separately. Once pretraining is completed, an end-to-end fine tuning is performed.

Dynencoder is shown to be successful in synthesizing dynamic textures. One can think of a Dynencoder as a compact way of representing the spatiotemporal information of a video. As such, the reconstruction error of a video given a Dynencoder can be used as a mean for classification.

\subsubsection{LSTM Autoencoder Model}

\begin{figure}[t!]
  \centering
  \includegraphics[width=0.85\textwidth]{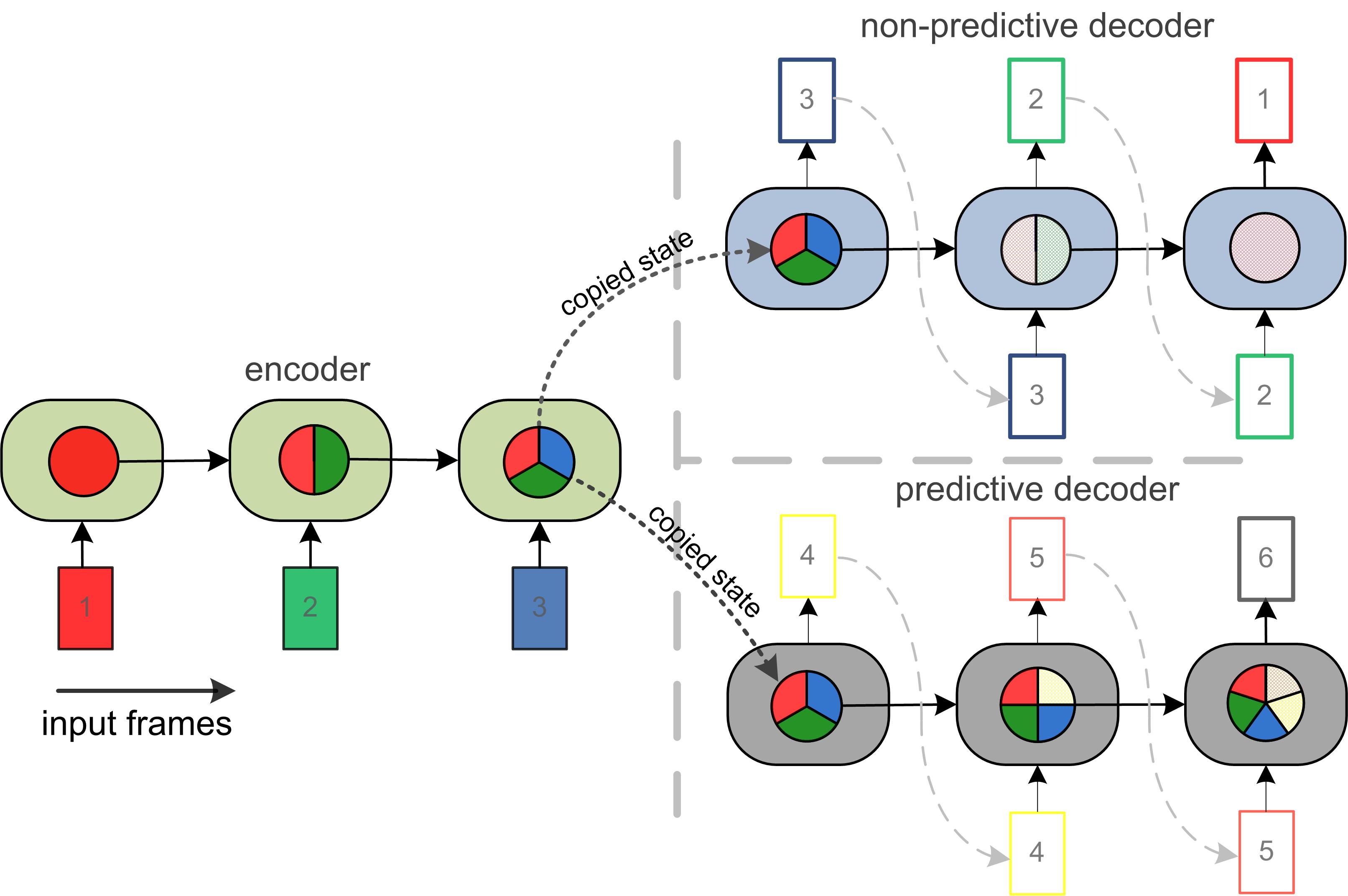}
  \caption{The composite generative LSTM model by ~\cite{Srivastava_LSTM15}. The internal states (represented by the circle inside) of the encoder LSTM captures a compressed 
version of the input sequence (\eg, frames $1,2$ and $3$). The states thereafter are copied into two decoder models, which are reconstructive and predictive. The reconstruction decoder attempts to reconstruct original frames in the reverse order. The predictive model is trained on predicting the future frames $4,5$ and $6$. The colors on the state markers indicate the presence of information from a particular frame.}
  \label{fig:srivastavRNN}
\end{figure}

Generative models for action recognition are expected to discover long-term cues, making deep models with LSTM cells natural choices. 
To this end, \cite{Srivastava_LSTM15} introduced the LSTM autoencoder model as illustrated in Fig.~\ref{fig:srivastavRNN}. The LSTM autoencoder consists of two RNNs, namely the \emph{encoder} LSTM and the \emph{decoder} LSTM. The encoder LSTM accepts a sequence (as input) and learns the corresponding compact representation. The states of the encoder LSTM  contain the appearance and dynamics of the sequence. As such, the compact representation of a sequence is chosen to be the states of the encoder LSTM. The decoder LSTM receives the learned representation to reconstruct the input sequence. For more details, see Fig.~\ref{fig:srivastavRNN}.

The LSTM autoencoder can be used to predict the future of a sequence as well. In practice, a composite model that both reconstructs the input sequence and predicts its future delivers the most accurate responses.

\subsubsection{Adversarial Models}

To sidestep various difficulties in training deep generative models, \cite{Goodfellow_Advers14} introduce the adversarial networks where a generative model competes with a discriminative model known as an adversary. The discriminative model learns to determine whether a sample is coming from the generative model or the data itself. During training, the generative model learns to generate samples that share more similarities to the original data, 
while adversary model improves its judgments on whether a given sample is authentic or not. 
\cite{Mathie_MultiScale15} adopt the adversarial methodology to train a multi-scale convolutional network for video prediction. They exploit adversarial training to have convolutional networks that avoid pooling layers. They also provide a discussion on the advantages of pooling in generative models.

\subsection{Temporal Coherency Networks}

Before concluding this part, we would like to bring the notion of \emph{temporal coherency} into perspective. Temporal coherency is a form of weak supervision and states that consecutive video frames are correlated both semantically and dynamically (\ie, abrupt motions are less likely). {For actions, even stronger connections between spatial and temporal cues
exist~\citep{Rahmani_CVPR15}.}
A sequence is called coherent if its frames are in the correct temporal order. Temporal coherency can be learned by a deep model if the model is fed by ordered and disordered sequences as positive and negative samples, respectively. This concept has been used by \cite{Goroshin_SpatioTemp14} and \cite{WangGupta_ICCV15} to learn robust visual representations from unlabeled videos. 

\begin{figure}[!b]
  \centering
  \includegraphics[width=0.75\textwidth]{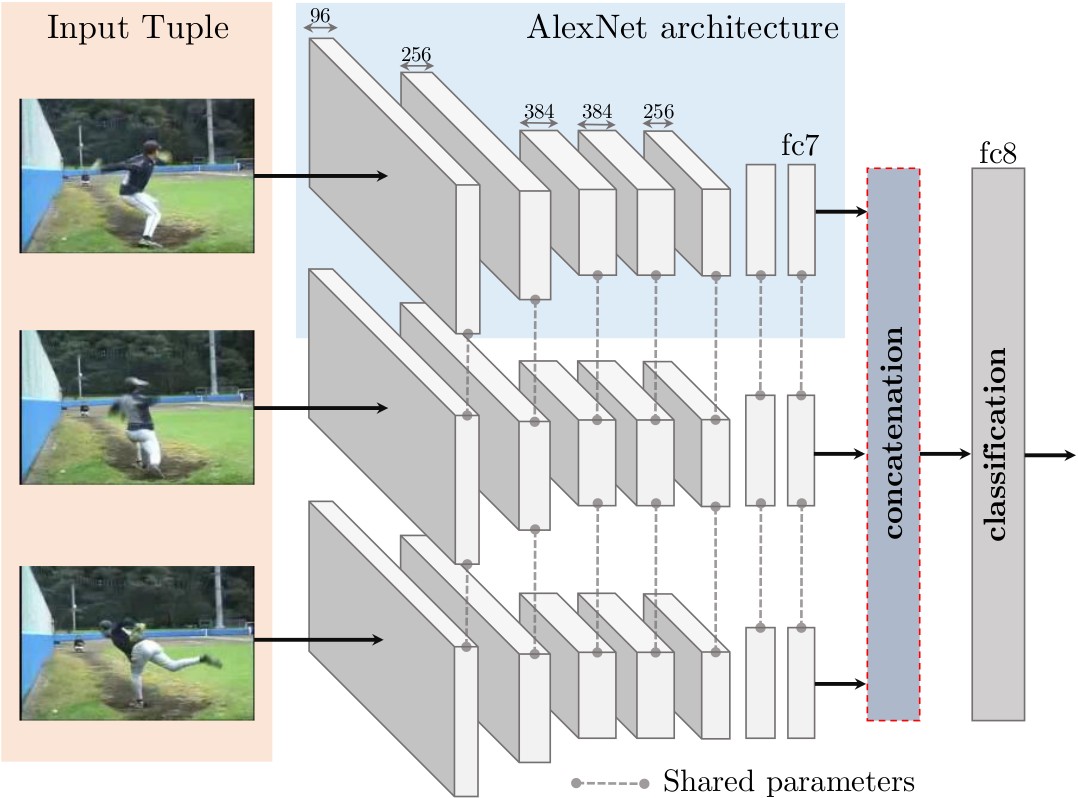}
  \caption{The Siamese Triplet network used by \cite{Misra_SeqVeri16}. 
  Each network is expected to capture the motion and pose of actions.}
  \label{fig:MisraSeq}
\end{figure}

\cite{Misra_SeqVeri16} study how temporal coherency can be used to train deep models for action recognition and pose estimation. In particular, a \emph{Siamese Network}~\citep{Chopra_CVPR05,Varior_ECCV16,Lu_ICME16} (see Fig.~\ref{fig:MisraSeq}) is trained with tuples to determine whether a given sequence is coherent or not. Empirically, it has been shown that 
\begin{itemize}[noitemsep]
\item Compared to other supervised pretrained methods, \eg, ImageNet ~\citep{ILSVRC15}, learning by tuples give more attention to human poses.

\item Selection of tuples in the frames with rich motions will avoid ambiguities between positive and negative tuples.

\item Compared to networks trained from scratch, pretrained networks based on the temporal coherency have potential to improve the accuracy.
\end{itemize}

{We note that the temporal coherency is not always a strong assumption to rely on. For example, an abrupt scene variation such as an advertisement shown during an sport event (\eg, in SPORTS-1M data) can violate the temporal coherency easily~\citep{Li_CVPR16}.}

\begin{figure}[!b]  
  \centering
  \includegraphics[width=0.75\textwidth]{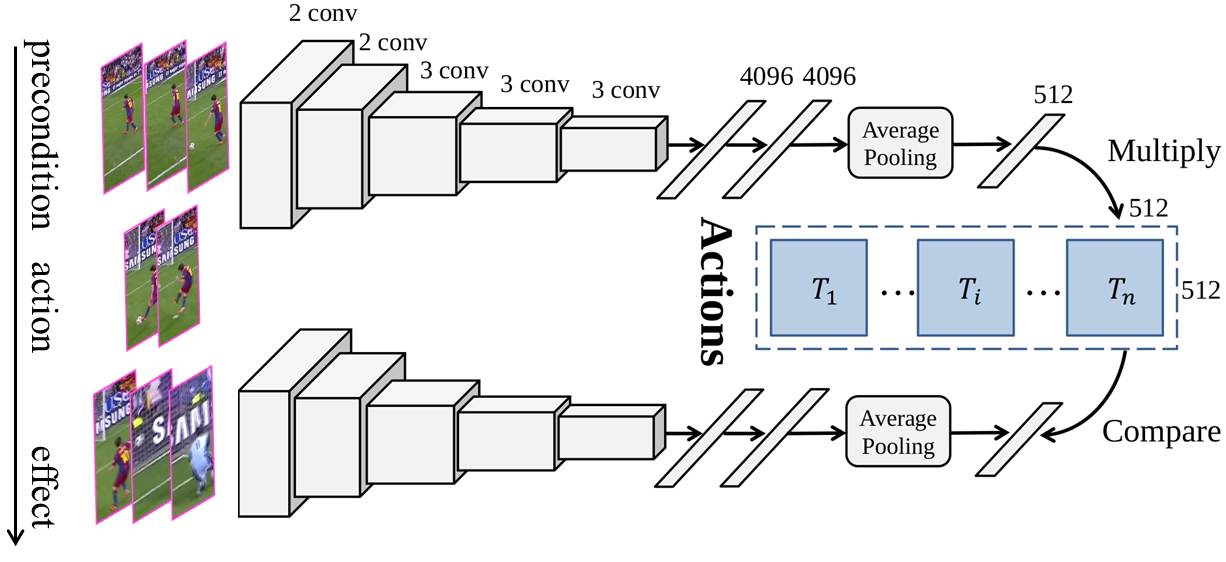}
    \caption{The parallel convolutional structures are used in extraction of precondition and post-effect features.}
    \label{fig:actionsTransformations}
\end{figure}

Another related study to temporal coherency is the work of \cite{WangGupta_ActTrans15} where an action is split into two phases
for classification. More specifically, a video with frames $\{\Vec{x}_1,\Vec{x}_2,\cdots,\Vec{x}_n\}$ is split into the precondition set $\Mat{X}_p = \{\Vec{x}_1,\Vec{x}_2,\cdots,\Vec{x}_p\}$ and effect set $\Mat{X}_e = \{\Vec{x}_e,\Vec{x}_{e+1},\cdots,\Vec{x}_n\}$. The cardinality of both sets are learned by the deep model. An action is then identified by the transformation required to map a high-level descriptor extracted from $\Mat{X}_p$ to a high-level descriptor extracted from $\Mat{X}_e$. In particular, the high-level descriptor and transformations are learned using the Siamese Networks  (see Fig.~\ref{fig:actionsTransformations} for details).

{Rank pooling ~\citep{Fernando_VidDarwin15} is an effective solution for capturing temporal evolution of a sequence. In its original form, learning of video representations (through ranking) and action  classification are done separately. This is due to the fact that, unlike other pooling operations such as max pooling, a closed form solution for the rank pooling operation is not readily available. Recently,~\cite{Fernando_ICML16} suggest an end-to-end learning scheme that learns both the pooling operation and the classifier with back propagation. A related work, though not being a deep learning based solution, is the \emph{hierarchical} rank pooling~\cite{Fernando_HRPCVPR16} that aims to encode multiple levels of dynamical granularity in videos by iteratively applying the rank pooling operations.}

For completeness, we conclude this section by discussing the work of~\cite{Ranzato_VideoM14}.
\cite{Ranzato_VideoM14} state that the success of language modeling through RNNs is a result of the discrete information space. Based on this, they introduce a discrete structure for video frames by quantizing them with a representative collection of image patches. 
Not surprisingly, natural videos seem to lack the dynamics word sequences possess, hinting why language models are superior to their video counterparts. Based on their observations, \cite{Ranzato_VideoM14} suggest training a recurrent convolutional network to predict long sequences may lead to better robustness in video modeling.

%===================================================================================================================================

{
\section{A Quantitative Analysis}

In this section we provide a high level analysis of the aforementioned solutions. We highlight the performance of some notable examples 
(see Table~\ref{tab:Performance_Tab}) in our discussion. Having our focus on performances, we also discuss challenges to be addressed and possible directions future solutions may take.
}

{
\subsection{What is measured by action datasets?}

Performance of the action recognition methods is usually compared on a few publicly available datasets. A comprehensive list of available datasets along their details is given in Table~\ref{tab:datasets}. In general, there is no such thing as a `` universal solution'', \ie, a solution  that can be applied to any given dataset. As such and if possible, we provide pointers as to why a solution is (un)successful in a scenario.}

Along the progress of solutions, action datasets are also evolved in terms of their complexities. The complexity of a dataset is usually described by the resemblance of its contents to reality. For example, the  KTH and the Weizmann datasets listed on the top of Table~\ref{tab:datasets} contain  human actions in controlled conditions (\eg, limited camera motion, almost zero background clutter). Furthermore, their scope is limited to basic actions such as walking, running and jumping. {Hence, comparing solutions on the KTH and Weizmann datasets is less insightful unless a specific need is considered. Having said this, we acknowledge that both datasets are useful if motion patterns are considered.
}

With the increasing complexity, we get to datasets that are composed of YouTube videos, movies and television broadcast snippets (\eg, HMDB-51, UCF-101). YouTube videos are mostly recorded by nonprofessionals with handycams. As a result, they contain camera motion (and shakes), viewpoint variations and resolution inconsistencies. {To perform well, a solution must compensate the aforementioned variations. One can observe that in the HMDB-51 and UCF-101 datasets, the actions are well cropped in the temporal domain. Therefore, these datasets are not well-suited for measuring the performance of action localization. An interesting feature of the HMDB-51 and UCF-101 datasets is the inclusion of what we would like to call subtle classes. Examples are \emph{chewing} and \emph{talking} or  \emph{playing violin} and \emph{playing cello}. Learning to distinguish between subtle classes requires a deeper understanding of spatial and temporal clues.} 

{Movies and many sports broadcasts are filmed from several viewpoints, and then edited into one stream. This brings sudden viewpoint variations to the video streams. Both Hollywood2 and Sports-1M datasets contain view-point/editing complexities. Furthermore, the actions usually occur in a small portion of the clip. To make the recognition more challenging, SPORTS-1M dataset also contains scenes of spectators and banner adverts. Therefore, methods that rely on temporal coherency may fail on Sports-1M. We note that both SPORTS-1M and Hollywood2 datasets are specifically annotated by text and script analysis albeit labeling is noisy~\citep{Laptev_Realistic08,KarpathyLarge14}.} 

As mentioned earlier, all action classes in the  HMDB-51,UCF-101, Hollywood2 and Sports-1M datasets \emph{cannot} be distinguished by motion clues. In such situations, the objects contributed to the actions become important as certain actions are defined by the related objects. A good example for this is the 23 distinct types of billiard categories given in Sports-1M. Hence, 
algorithms benefiting from object details are expected to perform better. 

{
Deep architectures are notorious for their data-hungry nature. As such, tuning deep networks on small and medium size action datasets such as KTH and Wiezmann is difficult and often leads to unsatisfactory performance~\citep{Sun_ICCV15}. The Sports-1M dataset is assembled to alleviate this limitation, making training and tuning very deep networks possible.}

\subsection{Recognition Results}

In Table~\ref{tab:Performance_Tab}, we provide a comprehensive list of 31 must-know methods along their accuracies on seven challenging action datasets. The accuracies are reported directly from the original works. 
Instead of considering each case individually, we opt to give a high level comparison between various classes of solutions. 

\subsubsection*{Almost equal performance?}

{
A quick look at the accuracy scores shows that the state-of-the-art solutions based on both representation and deep learning perform equally well. This would have been an unexpected observation if image classification is considered. For example, the stacked FV encodings of trajectory descriptors ~\citep{Peng_SFV14} outperforms the state-of-the-art deep learning based solutions ~\citep{Feichtenhofer_CVPR16,varol_LTC16} on HMDB-51  without trajectory decision fusion. In contrast, the gap between similar solutions (\eg, comparison between SIFT $+$ FV and CNNs in~\cite{krizhevsky_imagenet_2012}) is contradictory when it comes to image classification. Among various reasons one can think of, the insufficiency of data cannot be disregarded\footnote{While ImageNet contains over 1000 training instances per class, HMDB-51 has only around 70.}. A dominant theme to get around this limitation is to benefit from models pre-trained on images~\citep{varol_LTC16,Feichtenhofer_CVPR16,Simonyan_Two14}.}

\subsubsection*{State-of-the-art solutions}

\paragraph*{\textbf{Handcrafted Solutions}}%

{Focusing on the handcrafted solutions, a performance milestone is achieved by the introduction of \emph{dense trajectory descriptors}~\citep{Wang_ImpTraj13}. The descriptor can be easily incorporated in various pooling strategies such as \emph{FV}'s~\citep{Peng_SFV14} and \emph{Rank-Pooling} ~\citep{Fernando_VidDarwin15}, leading to competitive results on the  HMDB-51 and UCF-101 datasets. }

\paragraph*{\textbf{Deep-net Solutions}}%

{Turning our attention to deep solutions, we find that the spatiotemporal networks~\citep{KarpathyLarge14,Tran_C3D15, varol_LTC16} and two-stream networks~\citep{Simonyan_Two14, Feichtenhofer_CVPR16} outperform other network structures. Lately, both of these structures are equipped with 3D convolution filters. Examples include the use of 3D convolutions and pooling in \cite{Feichtenhofer_CVPR16} and \cite{varol_LTC16}. The work in \cite{Limin_GoodPrac15} also suggests deeper models help boosting the performances. However, training deeper networks demands more rigorous data augmentation techniques (\eg, temporal crops by random clip sampling, frame skipping).
}

{
\subsubsection*{Fusion with dense trajectories seems to always help.}

The recognition accuracy (\emph{see} Table ~\ref{tab:Performance_Tab}) of most the state-of-the-art deep learning based  solutions~\citep{Tran_C3D15,Feichtenhofer_CVPR16,varol_LTC16} can be improved based on the observations made in \cite{Wang_ImpTraj13} \footnote{In addition to the decision level fusion, the work of~\cite{Wang_TDD15,deSouza_ECCV16} suggests employing hybrid fusion methods.}. This indicates that the structures learned by deep networks are complementary to the \emph{handcrafted} trajectory descriptors. It is worth mentioning that both deep networks (in most cases) and trajectory descriptors consider similar inputs (\ie, RGB and optical flow frames). In~\cite{Simonyan_Two14}, it is observed that some filters in the temporal stream respond to spatial gradients. Similarly, the \emph{MBH} trajectory descriptor~\cite{Wang_ImpTraj13} is also derived using spatial gradients on the optical flow frames.
}

\subsection{What algorithmic changes to expect in future?}

{
Following the trend of other developments in computer vision, moving towards deep architectures for action recognition is dominating the action recognition research lately. 
Given the difficulty of training deep networks when it comes to video data, \emph{knowledge transfer}, \ie benefiting from models trained on images or other sources, is an avenue to explore. 
A related and less investigated problem for \emph{knowledge transfer} in deep networks is the idea of \emph{heterogeneous domain adaptation} ~\citep{Tsai_CVPR16,Hoffman_IJCV13}.

Considering deep architectures for action recognition, the keywords to remember would be 3D convolutions, temporal pooling, optical flow frames, and LSTMs. Though the aforementioned elements are developed individually, novel methods aim at blending them to boost the performance ~\citep{Donahue_LRCN15,varol_LTC16,Feichtenhofer_CVPR16}. We consider this might be an indication of convergence towards a generic form of deep architectures for spatiotemporal learning. 

Another point to remember is that, to boost the performance, carefully engineered approaches are needed. For instance, data augmentation techniques~\cite{Limin_GoodPrac15}, foveated architecture~\citep{KarpathyLarge14} and distinct frame sampling strategies~\citep{Simonyan_Two14,Feichtenhofer_CVPR16} have been shown to be essential. 

\subsection{Bringing action recognition into life}}

{Action recognition has advanced from recognition in controlled environments~\citep{Rohr_Model94,Bobik_MEI01,Blank_STShapes05} 
to solutions that target more realistic activities (see Table ~\ref{tab:Performance_Tab}). However, in order to use these solutions in real-life scenarios, deeper understanding in the following areas is required: 
\begin{itemize}

\item  Action recognition for a practical applications involves joint detection and recognition from a sequence. Some recent works address joint segmentation and recognition of actions~\citep{Carvajal_MLS16,Carvajal_14,Borzeshi_SPL13}. 

\item Rather than recognizing actions from a big pool of classes, constraining into a refined set of actions can be useful in practical applications (\eg, cooking activities of~\cite{Rohrbach_CVPR12} ). Therefore,  fine-grained action recognition tasks~\citep{Ni_CVPR16,Singh_CVPR16,Lea_ECCV16} that have been already receiving growing attention from the community, can shape 
the future solutions and associated problems.
\end{itemize}
}

\begin{sidewaystable}[h]
	\scriptsize
    \centering
    \caption{Datasets for action recognition}
    \label{tab:datasets}
\begin{tabular}{llllllll}
\hline
\multicolumn{1}{c}{Dataset} & \multicolumn{1}{c}{Source} & \multicolumn{1}{c}{No. of Videos} & \multicolumn{1}{c}{Video Duration} & \multicolumn{1}{c}{Training Protocol} & \multicolumn{1}{c}{No. of Classes} & \multicolumn{1}{c}{Videos/Class} & \multicolumn{1}{c}{Example Classes} \\ \\ \hline \\
KTH~\citep{Schuldt_BoW04} & \Three{Recoded videos on both outdoors and indoors} & \multicolumn{1}{c}{600} & ~4s & \Three{Training and Testing are divided on subjects} & \multicolumn{1}{c}{6} & - & \Two{Walk, Jog, Run} \\ \\ \hline \\
Weizmann~\citep{Blank_STShapes05} & \Three{Outdoor video recordings on still backgrounds} & \multicolumn{1}{c}{90} &  & \Three{Leave out one cross validation} & \multicolumn{1}{c}{10} & - & \Two{Walk, Jump, Jumping Jack, Skip}\\ \\ \hline \\
UCF-Sports~\citep{Rodriguez_mach08} & \Three{Television sports broadcasts(eg. BBC, ESPN) – (780x480)} & \multicolumn{1}{c}{150} & ~6.39s & \Three{Classification accuracy on provided train test splits by Tian Yan, Discriminative figure-centric models for} & \multicolumn{1}{c}{10} & 6 - 22 & \Two{Diving, Golf-swing, Kicking} \\ \\ \hline \\
Hollywood2~\citep{Marszalek_Hollywood209}& \Three{Clips from 69 Hollywood movies (33 training and 36 testing) annotated based on movie script} & \multicolumn{1}{c}{1707} &  & \Three{mAP of each class(884 Test videos and 823 training videos obtained from separate training and testing movies)} & \multicolumn{1}{c}{12} & 20 - 140 & \Two{Answer-phone, Eat, Handshake} \\ \\ \hline \\
Olympic Sports~\citep{Niebles_TempStruct10} & \Three{Youtube Video Sequences} &  &  & \Three{mAP of each class on provided train-test splits} & \multicolumn{1}{c}{16} & ~50 & \Two{High-jump, Long-jump, Tripple-jump} \\ \\ \hline \\
HMDB-51~\citep{Kuehne2013} & \Three{Youtube, Movies} & \multicolumn{1}{c}{7000} &  2 - 3s & \Three{Classification accuracy of 30 test clips with training on 70 clips (3 splits are provided)} & \multicolumn{1}{c}{51} & Over 101 & \Two{Brush-hair, Kick, Kiss} \\ \\ \hline \\
UCF-50~\citep{Reddy_UCF5013} & \Three{Youtube Video Sequences} & \multicolumn{1}{c}{-} & - & \Three{Leave out one cross validation} & \multicolumn{1}{c}{50} &  - & \Two{Rowing, Fencing, Punch} \\ \\ \hline \\
UCF-101~\citep{Soomro_UCF101} & \Three{Youtube Video Sequences} & \multicolumn{1}{c}{13320} & 2 - 5 s & \Three{Classification accuracy on 3 train and test spits} & \multicolumn{1}{c}{101} & Over 100 & \Two{Diving, Skiing, Apply Eye Makeup} \\ \\ \hline \\
Sports 1-M~\citep{KarpathyLarge14}& \Three{Youtube Sports Videos annotated automatically from YouTube topics} & \multicolumn{1}{c}{1133158} &  & \Three{70\% of as training while testing and validation sets are respectively 20\% and 10\%.} & \multicolumn{1}{c}{487} & 1000 - 3000 & \Two{Cricket, disc golf, gliding} \\ \\ \hline \\
\end{tabular}
\end{sidewaystable}

\begin{table}[tb]
\centering
\caption{Accuracy of action recognition techniques (Numbers are true recognition accuracy given in percentages. * Datasets in which the mean average precision is reported). The column \emph{Type} indicates whether a method is purely Deep-net based(D), Representation Based(R) or Fused Solution(F).}
\begin{adjustbox}{max width=1.08\textwidth, center, max height=10cm}
\label{tab:Performance_Tab}
\begin{tabular}{ | L{5cm} | L{7.2cm} |C{1cm}| R{1.5cm} | R{1.4cm} | R{1.4cm} | R{1.4cm} | R{1.4cm} | R{1.4cm} | R{1.4cm} |}
\hline
\multirow{3}{*}{Reported Paper} & & & \multicolumn{7}{c|}{\multirow{2}{*}{Dataset}} \\
 & Method & Type &\multicolumn{7}{c|}{} \\ \cline{4-10} 
 & & & HMDB51 & UCF101 & UCF50 & UCF-Sports* & Hollywood2* & Olympic Sports* & Sports1M \\ \hline
%--------------------------------------------------------------
~\cite{Wang_Traj11} & Dense Traj (Traj + HOG+HOF+MBH) & R &  &  &  & 88.2 & 58.3 &  &  \\ \hline
%-------------------------------------------------------------
~\cite{Kliper-Gross12} & Motion Interchange Patterns & R & 29.2 & & 68.5 &  &  &  &  \\ \hline
%--------------------------------------------------------------
                              & General & &26.9 &  &  &  &  &  &  \\
~\cite{Sadanand_ActionBank12} & Video Wise & R & &  & 76.4 &  &  &  &  \\
                              & Group Wise & & &  & 57.9 &  &  &  &  \\ \hline
%--------------------------------------------------------------
~\cite{Oneta_FV13} & MBH + SIFT + Sqrt + L2 Normalization & R &54.8 &  & 90 &  & 63.3 & 82.1 &  \\ \hline
%--------------------------------------------------------------
 & Without Human Detector & R & 55.9 &  & 90.5 &  & 63 & 90.2 &  \\
~\cite{Wang_ImpTraj13} & With Human Detector & & 57.2 &  & 91.2 &  & 64.3 & 91.1 &  \\ \hline
 %--------------------------------------------------------------
~\cite{Jain_DCS13} &Traj + HoG + HoF + MBH + DCS on $w$-flow & R & 52.1 &  &  &  & 62.5 &  &  \\ \hline
%-------------------------------------------------------------
~\cite{Peng_SFV14} & Stacked FVs + FV & R & 66.8 &  &  &  &  &  &  \\ \hline
%-------------------------------------------------------------
~\cite{Peng_GoodPractices14} & Hybrid-BoW & R &61.1 & 87.9 & 92.3 &  &  & &  \\ \hline
 %-------------------------------------------------------------
~\cite{Kantorov_MPEGFlow14} & MPEG-Flow : VLAD encodings of & R &46.3 &  &  &  &  &  &  \\ \hline
%--------------------------------------------------------------
~\cite{Gaidon_Cluster14} & SDT tree ATEP & R &41.3 &  &  &  & 54.4 & 85.5 &  \\ \hline 
%--------------------------------------------------------------
~\cite{Simonyan_Two14} & Two-stream (CNN-M-2048) & D & 59.4 & 88.0 &  &  &  &  &  \\ \hline
%--------------------------------------------------------------
 & Transfer Learning on Sports 1M& & & 65.4 &  &  &  &  &  \\
~\cite{KarpathyLarge14} & Clip Hit @ 1 - Slow Fusion &D &  &  &  &  &  &  & 41.9 \\
 & Video Hit @ 1 - Slow Fusion & &  &  &  &  &  &  & 60.9 \\ \hline
%--------------------------------------------------------------
~\cite{Sun_ICCV15} & Factorized Spatio Temporal Conv. Nets&D & 59.1 & 88.1 &  &  &  &  &  \\ \hline
%--------------------------------------------------------------
					    & Two-Stream (ClarifaiNet)& & & 88.0 &  &  &  &  &  \\
~\cite{Limin_GoodPrac15} & Two-Stream (GoogLeNet)&D & & 89. 3&  &  &  &  &  \\
					    & Two-Stream (VGG-16)& & & 91.4 &  &  &  &  &  \\ \hline
%--------------------------------------------------------------
~\cite{Wang_TDD15} & TDD + ~\cite{Wang_ImpTraj13}& F & 65.9 & 91.5 &  &  &  &  &  \\
 & TDD (Only)& F & 63.2 & 90.3 &  &  &  &  &  \\ \hline
%--------------------------------------------------------------
 & Conv Pooling Hit@1 (Best)& &  &  &  &  &  &  & 72.4 \\
 ~\cite{Yue-Hei_Beyond15}& LSTM Hit@1 (Best)&D&  &  &  &  &  &  & 73.1 \\
 & Conv Pooling (Image + Opt Flow)& &  & 88.2 &  &  &  &  &  \\
 & LSTM (Image + Opt Flow)& &  & 88.6 &  &  &  &  &  \\ \hline
 %--------------------------------------------------------------
~\cite{Fernando_VidDarwin15} & Rank Pooling&R & 63.7 &  &  &  & 73.7 &  &  \\ \hline
 %--------------------------------------------------------------
~\cite{Donahue_LRCN15} & LRCN- Weighted Avnerage of RBG + Flow & R & & 82.9 &  &  &  &  &  \\ \hline
%--------------------------------------------------------------
~\cite{Wu_Fusing15} & Adaptive Multi-Stream Fusion &D & & 92.6 &  &  &  &  &  \\ \hline
%--------------------------------------------------------------
~\cite{Jiang_Unconst15} & TrajShape+TrajMF&R &48.4 & 78.5 &  &  & 55.2 & 80.6 &  \\
 & TrajShape+TrajMF+~\cite{Wang_ImpTraj13} & &57.3 & 87.2 &  &  & 65.4 & 91 &  \\ \hline
%-------------------------------------------------------------
~\cite{Lan_BeyondGauPyr12}& Multi-Skip Feat. Stacking &R &65.1 & 89.1 & 94.4 &  & 68.0 & 91.4 &  \\ \hline
%--------------------------------------------------------------
~\cite{Hoai_MIL15} & Proposed SSD + RCS &R &62.2 &  &  &  & 72.7 &  &  \\ \hline
%--------------------------------------------------------------
~\cite{Tran_C3D15} & C3D on SVM &D && 85.2 &  &  & &  &  \\ 
& C3D + ~\cite{Wang_ImpTraj13} on SVM& F && 90.4 &  &  & &  &  \\\hline
%--------------------------------------------------------------
~\cite{Misra_SeqVeri16} & ImageNet  pretrain + tuple verification&D & 29.9 &  &  &  &  &  &  \\
 & HMDB + UCF101 Labels Only& & 30.6 &  &  &  &  &  &  \\ \hline
%--------------------------------------------------------------
~\cite{WangGupta_ActTrans15}& Proposed Only (RBG + Opt Flow Networks)&D & 62 & 92.4 &  &  &  &  &  \\  \hline
%---------------------------------------------------------------
~\cite{Fernando_ICML16} & End to End Rank-pooling&D &  &  &  & 87 & 40.6 &  &  \\ \hline
%---------------------------------------------------------------
~\cite{Fernando_HRPCVPR16} & Hierarchical Rank-pooling (CNN Features)&D & 47.5  & 78.8 & &  & 56.8 &  &  \\ 
& Hierarchical RP on CNN+ ~\cite{Fernando_VidDarwin15}&F & 65.0 & 90.7 & &  & 74.1 &  &  \\ \hline
%---------------------------------------------------------------
~\cite{Li_CVPR16} & VLAD$^3$ &F&   & 84.7 & &  & & 90.8 &  \\ 
& VLAD$^3$ + ~\cite{Wang_ImpTraj13}&F&    & 92.2 & &  & & 96.6 &  \\  \hline
%---------------------------------------------------------------
~\cite{varol_LTC16} & LTC$_{flow + RGB}$&D & 64.8 & 91.7 &  & &  &  &  \\
& LTC$_{flow + RGB}$ + ~\cite{Wang_ImpTraj13}&F & 67.2 & 92.7 &  & &  &  &  \\\hline
%---------------------------------------------------------------
~\cite{Feichtenhofer_CVPR16} & Two Stream Fusion (VGG-16)&D & 65.4 & 92.5 &  & & &  &  \\
& Two Stream Fusion (VGG-16) + ~\cite{Wang_ImpTraj13} &F &69.2 & 93.5 &  & & &  &  \\\hline

%--------------------------------------------------------------
~\cite{deSouza_ECCV16} & Hybrid fusion of ~\cite{Wang_ImpTraj13} $+$ Deep-nets&F & 70.4 & 92.5 &  &  & 72.6 &  &  \\ \hline

%---------------------------------------------------------------
\end{tabular}
\end{adjustbox}
\end{table}

\section{Conclusion}

{Despite having  similarities to \emph{static image analysis}, video data analysis is far more complicated. A successful video analytic solution not only needs to overcome variations such as scale, intra-class diversities and noise, but also has to analyze motion cues in videos. 

Human action recognition can be considered as the queen of video analysis problems due to its wide applications and the complexity of the motion patterns produced by articulated body movements. In this survey, we investigate several aspects of the existing solutions for action recognition. We first review methods based on the handcrafted representations, and then focus on solutions that benefit from deep architectures. We provide a comparative analysis of these two prevailing lines of research.}

%===================================================================================================================================

\section*{Acknowledgment} 
We would like to thank our reviewers for pointing out the ways we could improve this survey. Further, we would like to thank \emph{Dr. Anoop Cherian} and \emph{Dr. Basura Fernando} for fruitful discussions and encouragement comments given for this work.

%===================================================================================================================================

\clearpage

\bibliography{master_bib}
\end{document}